\documentclass[a4paper,fleqn]{cas-sc}

\usepackage[authoryear]{natbib}
\usepackage{graphicx} 
\usepackage{algorithm}  
\usepackage{algpseudocode}
\usepackage{color}
\usepackage{setspace}
\usepackage{subcaption}
\usepackage{siunitx}
\usepackage{tabularx, booktabs, multirow}

\def\tsc#1{\csdef{#1}{\textsc{\lowercase{#1}}\xspace}}
\tsc{WGM}
\tsc{QE}
\tsc{EP}
\tsc{PMS}
\tsc{BEC}
\tsc{DE}

\usepackage{listings}
\usepackage[section]{placeins}
\usepackage{hyperref}

\usepackage{etoolbox}
\makeatletter
\patchcmd{\ttlh@hang}{\parindent\z@}{\parindent\z@\leavevmode}{}{}
\patchcmd{\ttlh@hang}{\noindent}{}{}{}

\makeatother
\usepackage{amsfonts}
\usepackage{bm}
\usepackage{bbm}
\usepackage{chngcntr}
\usepackage{indentfirst}
\usepackage{graphicx}

\usepackage{mwe}

\usepackage[export]{adjustbox}
\usepackage{graphicx}
\usepackage{caption}
\usepackage{subcaption}

\usepackage{booktabs,caption}
\usepackage[flushleft]{threeparttable}

\usepackage[font=small,skip=5pt]{caption}
\usepackage{multirow}
\usepackage[section]{placeins}
\usepackage[group-separator={,},group-minimum-digits=4]{siunitx}

\def\vc#1{\mathbf{\boldsymbol{#1}}}     
\def\tn#1{\boldsymbol{#1}}

\def\avg#1{\langle#1\rangle}
\def \D{{{\rm I\kern-.3em D}}}

\def\avg#1{\langle#1\rangle}

\def\vc#1{\mathbf{\boldsymbol{#1}}}     
\def\tn#1{\boldsymbol{#1}}

\def\avg#1{\langle#1\rangle}

\def\div{\operatorname{div}}
\def\grad{\nabla}

\def\vmu{\vc\mu}

\def\rmin{\underline r}
\def\rmax{\overline r}

\begin{document}
\let\WriteBookmarks\relax
\def\floatpagepagefraction{1}
\def\textpagefraction{.001}
\shorttitle{Convolutional Surrogate for 3D Discrete Fracture-Matrix Tensor Upscaling}

\shortauthors{M. Špetlík, J. Březina}

\title [mode = title]{Convolutional Surrogate for 3D Discrete Fracture-Matrix Tensor Upscaling}

\author[1]{Martin Špetlík}[type=editor,
                        auid=000,bioid=1,orcid=0000-0002-5310-2663]
\credit{Writing – original draft, Software, Experimentation}

\author[1]{Jan Březina}
\credit{Supervision, Writing – review \& editing, Software}

\address[1]{Institute of New Technologies and Applied Informatics, Faculty of Mechatronics, Informatics and
Interdisciplinary Studies, Technical University of Liberec, Studentská 1402/2, 461 17 Liberec, Czech Republic}

\begin{abstract}
Modeling groundwater flow in three-dimensional fractured crystalline media requires capturing the spatial heterogeneity introduced by fractures. Direct numerical simulations using fine-scale discrete fracture-matrix (DFM) models are computationally demanding, particularly when repeated evaluations are needed. 
We aim to use a multilevel Monte Carlo (MLMC) method in the future to reduce computational cost while retaining accuracy. When transitioning between accuracy levels, numerical homogenization is used to upscale the impact of the hydraulic conductivity of sub-resolution fractures.
To reduce the computational cost of conventional 3D numerical homogenization, we develop a surrogate model that predicts the equivalent hydraulic conductivity tensor, $\tn K^{eq}$, from a voxelized 3D domain representing a tensor-valued random field of matrix and fracture hydraulic conductivities. Fracture properties, including size, orientation, and aperture, are sampled from distributions informed by natural observations.
The surrogate architecture combines a 3D convolutional neural network with feed-forward layers to capture both local spatial patterns and global interactions. Three surrogates are trained on data generated by discrete fracture-matrix (DFM) simulations, each corresponding to a different fracture-to-matrix conductivity ratio. Their performance is evaluated across varying fracture network parameters and correlation lengths of the matrix field.
The trained surrogates achieve high prediction accuracy ($\text{NRMSE} < 0.22$) in a wide range of test scenarios. To demonstrate practical applicability, we compare conductivities upscaled by numerical homogenization and by our surrogates in two macro-scale problems: computation of equivalent tensors of hydraulic conductivity and prediction of outflow from a constrained 3D area.
In both cases, the surrogate-based approach preserves accuracy while substantially reducing computational cost.
Surrogate-based upscaling achieves speedups exceeding $100\times$ when inference is performed on a GPU.
\end{abstract}


\maketitle 

\printcredits

\doublespacing

\section{Introduction}
Understanding groundwater flow in fractured rock (\cite{Banks2002}) is essential for assessing the long-term safety of deep geological repositories for radioactive waste disposal, as noted in \cite{IAEA_2011_SSG14}. However, direct numerical simulation (DNS) of such highly heterogeneous media is usually computationally infeasible, especially when the flow is coupled with thermal, chemical, or mechanical processes. 
A common remedy is to upscale the numerous small-scale fractures into an equivalent hydraulic conductivity tensor. A comprehensive review of hydraulic conductivity upscaling techniques is provided in \cite{WEN1996ix}.
Yet, the fractal nature of fracture networks, discussed in detail by \cite{{Sahimi20110420}}, leaves no distinct scale-separation threshold to guide that upscaling.

The discrete fracture–matrix (DFM) framework addresses this challenge by coupling an explicit discrete fracture network (DFN) with a surrounding continuum representation (see \cite{Berre2019} for various conceptual models of fractured rock). This hybrid description enables the imposition of an arbitrary fracture-size cutoff, allowing for a smooth transition from full DNS to coarser, continuum-only models. When paired with fast and robust numerical homogenization (\cite{Auriault20090101}), the resulting coarse model offers a low-cost, low-fidelity approximation - an essential tool for overcoming the subsurface data gap through inverse analysis and rigorous uncertainty quantification.

Our primary motivation is the multilevel Monte Carlo (MLMC) method, a computationally efficient strategy for uncertainty propagation and sensitivity analysis. MLMC relies on constructing a hierarchy of correlated fine- and coarse-grid solution pairs, with further details provided in \cite{Giles2015}. 
However, applying MLMC directly to DFM models presents a significant challenge: the coarse model lacks the resolution needed to represent small-scale fractures. To capture their impact, an efficient yet accurate homogenization of their hydraulic conductivity contribution into the matrix properties of the coarse DFM model is required.

In our previous work \cite{Spetlik2024}, we introduced deep learning-based surrogates to accelerate the homogenization process for 2D DFM models with non-constant hydraulic conductivities in both the matrix and fractures.
In this paper, we extend the 2D framework to the more realistic 3D setting. The transition introduces additional challenges, including the increased geometric complexity of fractures and the higher dimensionality of both inputs and outputs for the surrogate models. 
The generation of tensor-valued spatial random fields (SRFs) representing matrix hydraulic conductivity at the micro-scale DFM models must be extended to three dimensions. Moreover, fracture properties are informed by natural data, such as the distribution of fracture orientations, which were previously assumed to be uniform in the 2D case study. To accommodate the increased complexity of 3D data, the surrogate model architecture, including its hyperparameters, requires adjustments.

The following review focuses specifically on studies relevant to 3D numerical homogenization and its neural network-based surrogates. For a broader overview, see \cite{Spetlik2024}.
To homogenize the hydraulic conductivity tensor from 3D DFM models, we build not only on our previous research but also on research conducted by~\cite{https://doi.org/10.1029/2001WR000756}. The authors homogenize a 3D cubic domain of a rock matrix containing planar fractures of a single characteristic size. Their model assumes homogeneous, isotropic matrix properties and distinct permeabilities for the matrix and fractures. Using tetrahedral meshes that explicitly resolve fractures, they compute the effective hydraulic conductivity by solving Darcy flow separately in each domain. Their work was later extended by~\cite{PhysRevE.76.036309} to account for power-law distributions in fracture sizes. Subsequently,~\cite{Lang20140815} investigate 3D DFM models and employ pressure gradient combined with flux averaging to compute equivalent hydraulic conductivity tensors on unstructured grids. \cite{Azizmohammadi2017} continue in this research direction and investigate the anisotropy and scale dependence of equivalent permeability tensors.

In parallel, deep learning techniques, particularly convolutional neural networks (CNNs), have been increasingly used to infer properties of rock media from geometric or statistical descriptions. 
CNNs process volumetric data efficiently and capture 3D spatial dependencies. \cite{Hong2020} train 3D CNNs to predict directional and mean permeabilities from binary images representing pore-scale geometries. More recently,~\cite{MENG2023104520} combine CNNs with transformer architectures to enhance feature extraction from 3D porous structures.
Applications of deep learning to 3D DFM models remain relatively scarce. Some studies, such as~\cite{RAO2020109850, VASILYEVA2021185, Wang2023}, address permeability prediction in realistic 3D scenarios, but primarily focus on pore-scale imaging rather than DFM representations. Others, like~\cite{STEPANOV2023114980, pr11020601}, employ feed-forward neural networks to predict equivalent permeability from micro-scale simulations, while~\cite{Cai_2023, FERREIRA2022104264} address the same overarching objective using graph-based and generative modeling strategies, albeit within slightly different problem formulations.
Although upscaling of DFM models using deep learning has been explored in~\cite{10.2118/203901-MS, Andrianov2022UpscalingOT}, these works assume constant hydraulic conductivities. In contrast, our approach targets surrogate modeling of equivalent conductivity tensors derived from fully resolved 3D DFM simulations with stochastic parameters, making it compatible with MLMC frameworks.

The remainder of this article is organized as follows. Section~\ref{dfm_model_section} introduces the discrete fracture-matrix (DFM) model and outlines the parameters used to define the discrete fracture network. It then describes the generation of spatial random fields (SRFs) for a matrix at a micro-scale, followed by an overview of test macro-scale problems. This section also presents the numerical homogenization procedure within the multiscale DFM framework.
Section~\ref{dataset_section} describes the datasets used for training the surrogate and the applied preprocessing methods. Section~\ref{surrogate_arch_section} details the architecture of the surrogate. The results are presented and discussed in Section~\ref{results_section}, while Section~\ref{conclussion_section} concludes the paper.

\section{Multiscale models of fractured media}\label{dfm_model_section}
This section first provides a stochastic description of fractured rock using
(i) a discrete fracture network (DFN) model and
(ii) an equivalent continuous medium (ECM) representation with a spatially correlated hydraulic conductivity tensor field.
Next, we present their coupling through the DFM formulation.
Finally, we formulate two steady-state, macro-scale test problems and detail the numerical homogenization procedure.

\subsection{Discrete fracture network}
We consider a spatially uncorrelated discrete fracture network (DFN) described by the distributions of fracture size, aperture, orientation, and density. Fractures are assumed to be square-shaped, with size $f_s$ drawn from a power-law distribution (\cite{Bonnet2001ScalingOF}):
\begin{equation}
    f_s \sim C r^{-\alpha}, \qquad
    C = \frac{1-\alpha}{\rmax^{1-\alpha} - \rmin^{1-\alpha}},
\end{equation}
where $\alpha$ is the power-law exponent, and $\rmin$ and $\rmax$ denote the fracture size bounds. The effective fracture aperture is considered a linear function of the size: $\delta = a f_s$. 
Each fracture is assigned an isotropic hydraulic conductivity according to the cubic law:
\begin{equation}
    k_f = \frac{g \rho_w}{12 \mu} \delta^2,
\end{equation}
with gravitational acceleration $g$, water density $\rho_w$, and dynamic viscosity $\mu$.  
Fracture orientations follow the Fisher distribution (\cite{Adler1999}), parameterized by trend (horizontal projection angle), plunge (angle from the horizontal plane), and concentration. Fractures are spatially uncorrelated, placed using a Poisson process. The fracture density is controlled via $P_{30}$ - the number of fractures per unit volume.

\subsection{Spatially correlated random field on the matrix}\label{matrix_srf_section}
To describe the impact of fractures that are beyond the resolution of a given model, we adopt the equivalent continuous medium (ECM) approach (\cite{Hadgu_comparative_2017, Kottwitz_Investigating_2021}). Hydraulic properties are represented by a random hydraulic conductivity tensor field $\tn K(x)$, which can be defined implicitly by homogenizing a DFN realization (see Section~\ref{sec:DFM}) or explicitly through its marginal (pointwise) distribution and correlation structure.
In the latter case, we represent the conductivity tensor field as:
\begin{equation}
  \tn K = \tn Q^T \tn \Lambda \tn Q,
  \qquad
  \tn \Lambda = \mbox{diag}(k_x, k_y, k_z).
\end{equation}
We assume isotropical unresolved fractures modeled by the uniformly distributed rotation matrix 
$\tn Q=\tn Q_\vc{N}\tn Q_\vc{R}$.
To impose rotations with a correlation structure, we use normalized vector fields $\vc R = \vc {\tilde R} / \|\vc {\tilde R}\|_2$ and $\vc  N = \vc{\tilde N} / \| \vc{\tilde N}\|_2$, where both $\vc {\tilde R} = (\tilde R_x, \tilde R_y)$ and $\vc{\tilde N} = (\tilde N_x, \tilde N_y, \tilde N_z)$ are composed of independent Gaussian fields with zero mean, unit variance, and a specified correlation length $\lambda$. The vector field $\vc{N}$ determines the rotation of the $z$-axis, while $\vc{R}$ specifies the azimuthal orientation of the $x$-axis in the $xy$-plane.

The principal hydraulic conductivities are given by the Gaussian vector 
$(k_x, k_y, k_z) = \exp \left( \vc\mu + \sqrt{\vc \Sigma}\,\vc{\tilde k}\right)$,
where $\vc\mu$ is the mean vector and $\vc \Sigma$ is the covariance matrix.
Consequently, the tensor field is defined by eight independent scalar fields: $\tilde R_x$, $\tilde R_y$, $\tilde N_x$, $\tilde N_y$, $\tilde N_z$, $\tilde k_x$, $\tilde k_y$, and $\tilde k_z$, each following an $N(0,1)$ distribution with Gaussian correlation of length $\lambda$. These fields are sampled using the GSTools library (\cite{GSTools}).

\subsection{Discrete fracture-matrix model}\label{sec:DFM}
The discrete fracture-matrix (DFM) approach couples a DFN with a surrounding rock matrix represented by a heterogeneous ECM conductivity field.
We shall provide a minimal formulation; for details and related concepts, see  \cite{Sandve_efficient_2012a, Berrone_Simulations_2013, Brezina2016Analysis}. 
In 3D, the matrix occupies a volumetric domain $\Omega_m$, while fractures are represented as embedded 2D surfaces $\Omega_f$, forming a mixed-dimensional mesh combining 3D and 2D finite elements.
Within $\Omega_m$, we consider the Darcy flow equation:
\begin{equation}
\div \delta_m\vc u_m = 0,\quad  \vc u_m = - \tn K_m \grad h_m,
\end{equation}
where $\vc u_m(\vc x)$ denotes the Darcy velocity $[m/s]$ and $\tn K_m(\vc x)$ is the hydraulic conductivity tensor $[m/s]$. The principal unknown is the piezometric head $h_m$ $[m]$. For consistency with the fracture domain equations, we set $\delta_m = 1$.
The corresponding equation on $\Omega_f$ is given by:
\begin{equation}
\div \delta_f \vc u_f = q^+ + q^-, \quad
\vc u_f =-k_f \grad h_f,
\end{equation}
where $\delta_f(\vc x)=a r_i$ $[m]$ is the aperture of fracture $i$, and $k_f(\vc x)=k_f^i$ denotes the isotropic hydraulic conductivity of the fracture fill. The sources $q^\pm$ denote outflows from the two aligned boundaries of $\Omega_m$ with normal vectors $\vc n^\pm$, coupling the fracture and matrix domain through Robin-like boundary conditions:
\begin{equation}
-\tn K_m \grad h_m \cdot \vc n^\pm
= q^\pm := \frac{k_f}{2\delta}\big (h_m^\pm - h_f\big) \qquad\text{ on }\Omega_f.
\end{equation}
The system is solved using the \texttt{Flow123d} simulator~(\cite{flow123d}), based on a mixed finite element method.

We now introduce two benchmark problems designed to evaluate the performance of the proposed homogenization surrogates. The second benchmark additionally serves as the micro-scale homogenization problem that will later be approximated by a surrogate.
The multiscale nature of this approach is further elaborated in the subsequent section.

\subsubsection{Constraint problem}\label{constraint_problem_3D}
We examine fluid flow through a cubic domain $\Omega = (0, L)^3$. 
Dirichlet boundary conditions are imposed by setting the pressure head $h = H$ at $x=0$ and $h = 0$ at $x=L$. No-flow boundary conditions are applied on the remaining boundaries ($Y$ and $Z$ faces).

\subsubsection{Anisotropy problem}\label{anisotropy_problem_3D}
To obtain the full hydraulic conductivity tensor, we solve three problems on $\Omega = (0, L)^3$, applying the boundary pressure head $P^1=x$, $P^2=y$, and $P^3=z$, respectively.
Mixed formulation with $RT_0$ finite elements provides the velocity vector $\vc u(e, P^j)$ on element $e$ for boundary condition $P^j$, $j=1, 2, 3$. We get the corresponding pressure gradient  $\nabla h(e, P) = -\tn K_e^{-1} \vc u(e, P)$. For a vector quantity $\vc v(e, P)$ on elements, we introduce a weighted average:
\begin{equation}
\avg{\vc v}^j = \frac{\sum_{e\in\mathcal T} |e| \delta_e\vc v(e, P^j)}{\sum_{e\in\mathcal T} |e| \delta_e}, \quad j=1, 2, 3
\end{equation}
With this notation, we are able to write down the least squares problem for the components of the symmetric equivalent tensor written in Voigt notation:

$\tn K^{eq} = (k_{xx}, k_{yy}, k_{zz}, k_{yz}, k_{xz}, k_{xy})$:
\[
-\begin{bmatrix}
    \tn A^1\\
    \tn A^2\\
    \tn A^3
\end{bmatrix}
(\tn K^{eq})^T
=
\begin{bmatrix}
\avg{\vc u}^1\\
\avg{\vc u}^2\\
\avg{\vc u}^3
\end{bmatrix},
\]
with blocks:
\[
A^i = 
\begin{bmatrix}
\avg{\nabla h}^j_x & 0 & 0 & 0 & \avg{\nabla h}^j_z & \avg{\nabla h}^j_y \\
0 & \avg{\nabla h}^j_y & 0 & \avg{\nabla h}^j_z & 0 & \avg{\nabla h}^j_x \\
0 & 0 & \avg{\nabla h}^j_z & \avg{\nabla h}^j_y & \avg{\nabla h}^j_x & 0  
\end{bmatrix}.
\]
\subsection{Numerical homogenization in the context of multiscale models}\label{num_hom_multiscale}
A single level of MLMC consists of a pair of correlated fine and coarse resolution models. To this end, we adopt a two-scale strategy where the coarse model is a macro-scale model using equivalent hydraulic conductivity tensors resulting from block numerical homogenization of micro-scale models. 

\begin{figure}[pos=t]
  \centering
  \includegraphics[width=\linewidth]{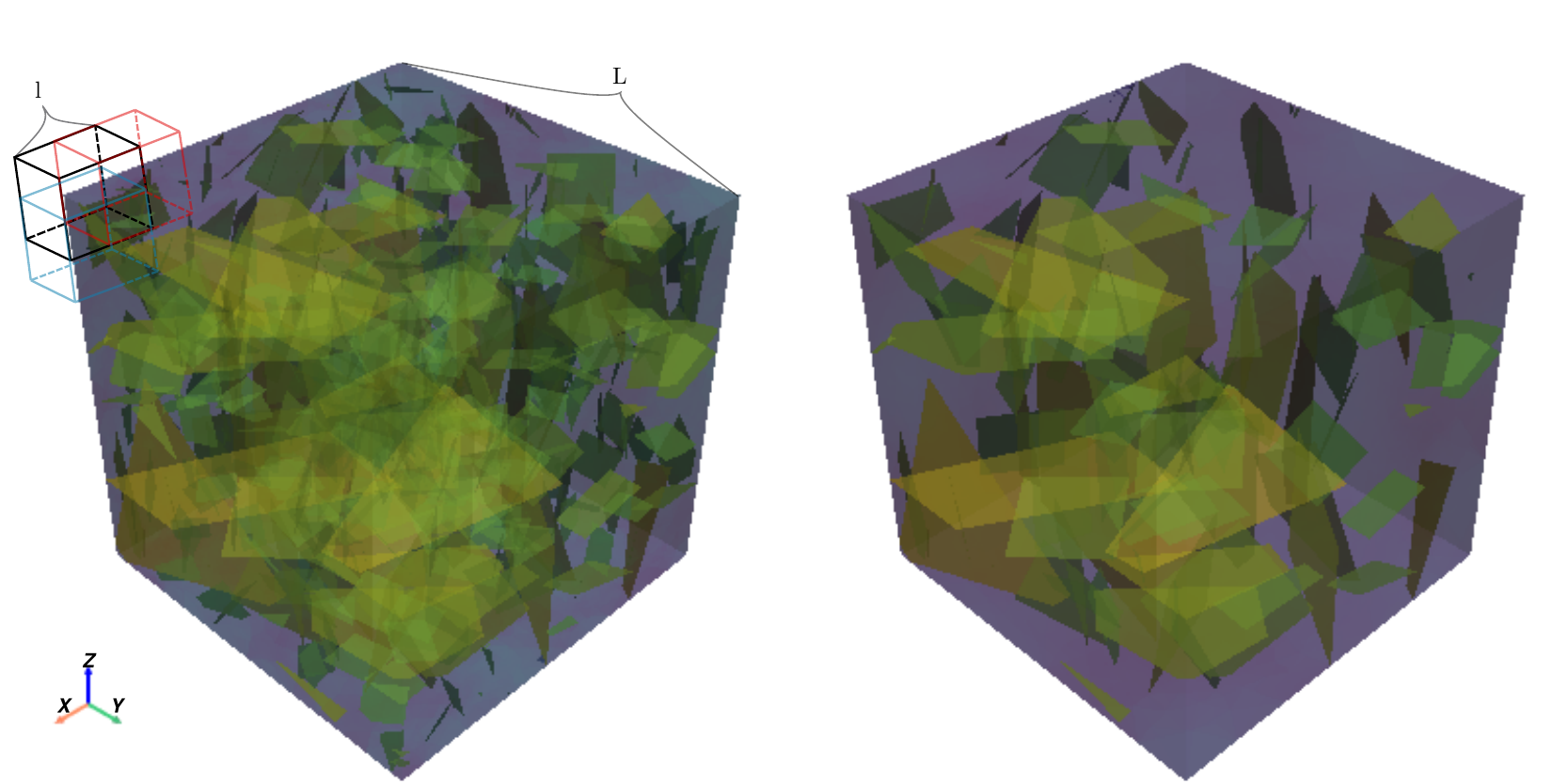}
  \centering
 \caption{Comparison of the fine DFM model fracture network $\mathcal{F}_{h, L}$, with $h < 5$ (left), and the corresponding coarse DFM model fracture network $\mathcal{F}_{H, L}$, with $H < 10$ (right). The illustration also shows the homogenization blocks of size $l = 15$ and their overlap $l/2$.}
 \label{fig:fine_coarse_fractures}
\end{figure}

\begin{figure}[pos=t]
  \centering
  \includegraphics[width=\linewidth]{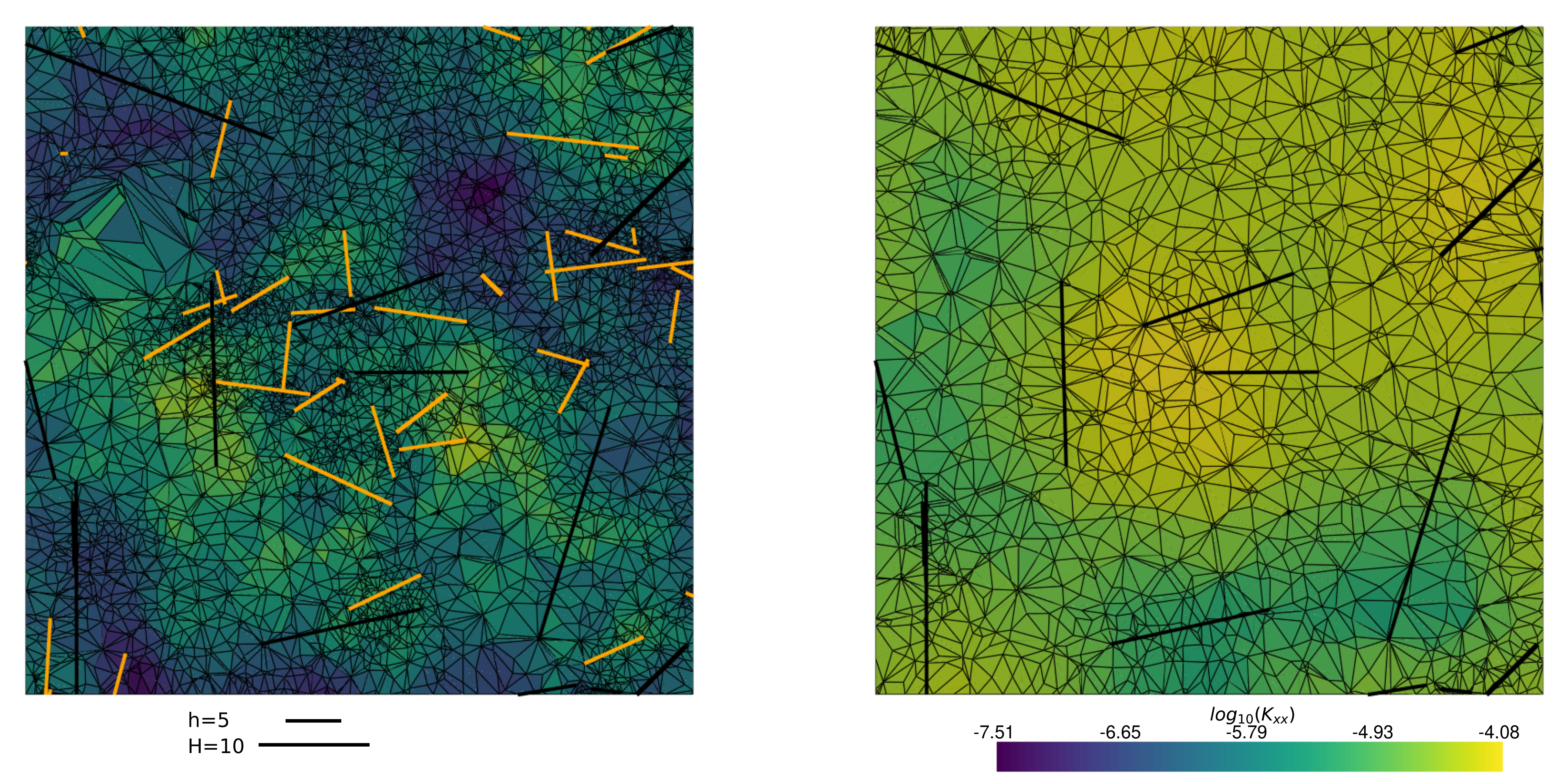}
  \centering
  \caption{{\bf Left:} The fine DFM model ($h<5$). Orange fractures ($\mathcal F_{h, H}$) and the hydraulic conductivity tensor field ($\tn K_h$) are homogenized using overlapping square blocks of size $l=15$. {\bf Right:} Corresponding coarse DFM model ($H<10$) with homogenized hydraulic conductivity tensor component $K_{xx}$. Notice the reduced range in the upscaled field. 
  Black fractures ($\mathcal F_{H, L}$) are not affected by homogenization.
  Only the $K_{xx}$ component of the hydraulic conductivity tensor is shown.}
  \label{fig:fine_coarse_srf_dfm_model}
\end{figure}

We start from the fine model. Its mesh discretization uses elements of maximum size $h$ and can explicitly capture fractures $\mathcal F_{h, L}$ in the size range $(h, L)$. The effect of fractures smaller than $h$ is represented by a realization of a spatially correlated random field of hydraulic conductivity $\tn K_h$ (see Section \ref{matrix_srf_section}).
The coarse model is discretized with a larger element size $H$ ($H > h$), and explicitly resolves fractures $\mathcal F_{H, L}$ in the size range $(H, L)$.
The effect of smaller fractures $\mathcal F_{h, H}$ is incorporated in the hydraulic conductivity tensor field $\tn K_H$, which is determined by numerical homogenization in terms of the micro-scale instance of the Anisotropy problem.  
The following procedure is adopted:
\begin{enumerate}
    \item Center points $\vc{x}$ of homogenization blocks are determined within domain of size $(L, L, L)$. Homogenization blocks are of size $l\times l \times l$ with overlap $l/2$, where $l=1.5 H$.
    \item The fracture network and the hydraulic conductivity field $\tn K_h$ are generated on an enlarged domain of side $L+l$, enabling homogenization blocks to capture boundary regions.
    \item For each block, the intersection of the fractures $\mathcal F_{h, H}$ with the block is determined, a compatible mesh with a maximum element size of $h$ is constructed, and the field $\tn K_h$ is interpolated to the mesh.
    \item An equivalent hydraulic conductivity tensor $\tn K^{eq}$ is calculated for each block to form the field $\tn K_H(\vc x)= P_h(\vc x, \mathcal F_{h, H}, \tn K_h)$, which is subsequently linearly interpolated onto the unstructured coarse grid cells.
\end{enumerate}
Figure \ref{fig:fine_coarse_fractures} shows an example of $\mathcal{F}_{h, L}$ (left), illustrating the homogenization blocks of size $l = 15$ and their overlap, alongside the corresponding $\mathcal{F}_{H, L}$ (right), for $L=60$, $H=10$, and $h=5$. Large fractures from the surrounding volume that intersect the domain of interest can produce small fragments, which remain visible even on the coarse DFM mesh. 
To illustrate the homogenization effect, Figure~\ref{fig:fine_coarse_srf_dfm_model} presents a slice in the ZY-plane through the center of the cube.
The homogenized representation of $\mathcal{F}_{h, H}$ (left, shown in orange) exhibits reduced variability in hydraulic conductivity in the corresponding coarse model (right).

To ease compatibility with convolutional neural networks, 
$\tn K_h$ is interpolated from $\tn K_G$, which is a hydraulic conductivity SRF generated on a regular grid over an enlarged domain of size: $L + l$. 
This approach allows us to use $\tn K_G$ for voxelization to avoid costly mesh generation.
Extension of this homogenization scheme to the full MLMC is a topic of ongoing research.

\FloatBarrier

\FloatBarrier

\section{Datasets}\label{dataset_section}
Let ${\mathcal{D} = \{(\vc{X}_j, \tn K^{eq}_j)\}_{j=1}^{D}}$ be a dataset of independent and identically distributed samples.
Here $\vc X_j \in \mathbb{R}^{6 \times F}$ is a vector of input features (voxelized upper triangle of hydraulic conductivity tensors of bulk and fractures, $F$ - number of voxels) and $\tn K^{eq}_j \in \mathbb{R}^6$ is the vector of components: $k_{xx}$, $k_{yy}$, $k_{zz}$, $k_{yz}$, $k_{xz}$, and $k_{xy}$ of the corresponding $\tn K^{eq}$ obtained by numerical homogenization described in Section~\ref{num_hom_multiscale}. Individual homogenization samples are generated independently on a fixed domain $\Omega_o = (0, 15)^3$. The DFN is generated with parameters listed in Table~\ref{tab:train_DFN_parameters} - adopted from \cite[p.~23]{SKB-R-09-20}. The fracture density \(P_{30}\) (number of fractures per unit volume) is derived from \(k_r\) and \(P_{32}\). To achieve a target average \(P_{30}\) of the whole population, the parameter \(r_0\), representing the minimum admissible fracture size, is adjusted. The size distribution parameters are set as \(\alpha = 2\), \(\rmin = 5\), and \(\rmax = 100\), while the aperture parameter is \(a = \num{1e-4}\).

Similarly to our 2D study (\cite{Spetlik2024}), we examine various fracture-to-matrix hydraulic conductivity ratios $K_f / K_m \in \{\num{1e3}, \num{1e5}, \num{1e7}\}$, forming three datasets:
\begin{itemize}
    \item \textbf{Dataset $\mathcal{A}$}: $K_f / K_m = \num{1e3}$,
    \item \textbf{Dataset $\mathcal{B}$}: $K_f / K_m = \num{1e5}$,
    \item \textbf{Dataset $\mathcal{C}$}: $K_f / K_m = \num{1e7}$.
\end{itemize}
The $K_m$ field is generated with a covariance matrix
\[
  \vc \Sigma =
  \begin{bmatrix}
    0.25 & 0.2 & 0.2 \\
    0.2  & 0.25 & 0.2 \\
    0.2  & 0.2 & 0.25 
  \end{bmatrix},
\]
and dataset-specific log-scale mean vectors:
$\vmu = [-4.0, -3.8, -3.9]$ for Dataset $\mathcal{A}$,
$\vmu = [-6.0, -5.8, -5.9]$ for Dataset $\mathcal{B}$,
and $\vmu = [-8.0, -7.8, -7.9]$ for Dataset $\mathcal{C}$.

Each dataset contains $75{,}000$ samples, evenly split between $P_{30} = 0.001$ and $P_{30} = 0.0025$. To enable generalization across different matrix correlation lengths, each dataset includes equal representation of $\lambda \in \{0, 10, 25\}$.
Samples are split into $80\%$ training set $\mathcal{L}$ and $20\%$ test set $\mathcal{T}$. The training set is further split into $80\%$ training and $20\%$ validation samples ($\mathcal{V}$).
\begin{table}
\centering
\caption{DFN parameters}
\label{tab:train_DFN_parameters}
\begin{tabular}[t]{lcccc}
\hline
Fracture  & Orientation & Size model & $P_{32}$ (mean fracture  & $P_{30}$\\
set name  & (trend, plunge), & power-law  &  surface area)  & (fracture intensity) \\
  & concentration & $k_r$ & ($r_0=1$, $564$ m) & ($r_0=1$, $564$ m) \\
\hline
\hline
NS &  (292, 1) 17.8  & 2.50 & 0.094 & 0.0196 \\
NE & (326, 2) 14.3  & 2.70  & 0.163 & 0.0427\\
NW & (60, 6) 12.9  & 3.10 & 0.098  & 0.0348\\
EW & (15, 2) 14.0  & 3.10 & 0.039 & 0.0138\\
HZ & (5, 86) 15.2  & 2.38 & 0.141 & 0.0247\\
\hline
\end{tabular}
\end{table}
\FloatBarrier

In order to form these datasets of homogenization blocks, the original input hydraulic conductivities $\tn K_G$ prescribed on a regular grid and fractures $\mathcal{F}_{h,H}$ contained within each homogenization block are voxelized onto a regular grid with a defined number of voxels. The fracture network is projected onto this grid by computing geometric intersections between fractures and voxels. For voxels intersected by fractures, the local conductivity tensors are computed as weighted combinations of the fracture and matrix ($\tn{K}_G$) conductivities, with weights proportional to the intersection volume fractions. The entire voxelization procedure is performed using the Bgem library (\cite{bgem}).

\begin{figure}
 \centering
 \includegraphics[width=0.97\textwidth]{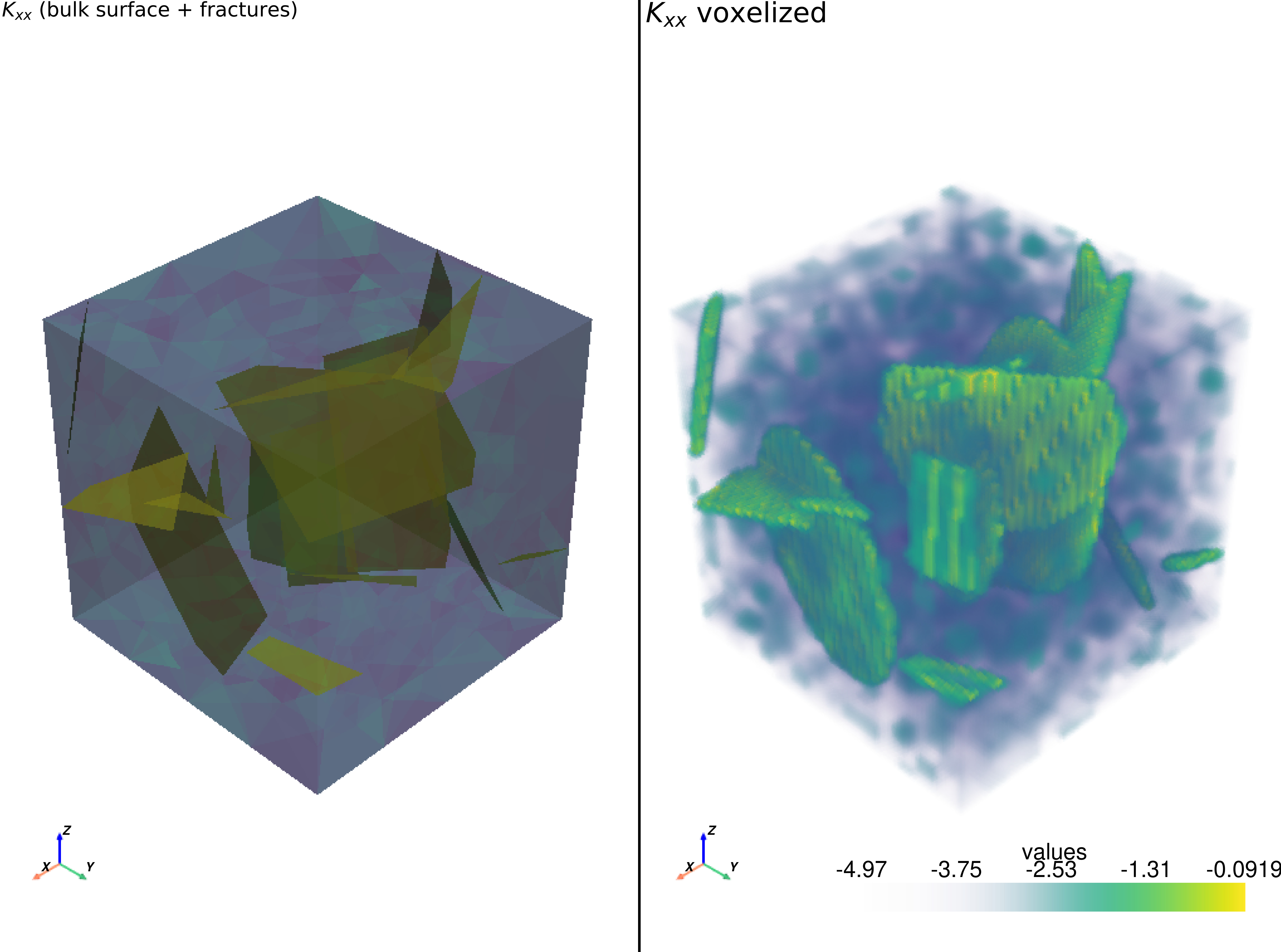}
 \caption{An illustration of an input hydraulic conductivity tensors ($K_{xx}$ component) on a mesh and its voxelized representation.}
 \label{fig:hom_sample_mesh_field_voxelized}
\end{figure}

Figure~\ref{fig:hom_sample_mesh_field_voxelized} shows an input mesh with fractures and its voxelized counterpart, illustrating the $K_{xx}$ component of the hydraulic conductivity tensors. The plots are adjusted to enhance fracture visibility.
As the voxelization quality required for accurate neural network predictions is not established, no systematic assessment was performed. To simplify and speed up, the voxelization algorithm considers disc-shaped fractures of equivalent area. According to our experiments, this approximation is able to capture key patterns and provides sufficient information to the surrogate model.

\FloatBarrier

\subsection{Datasets analysis}
\begin{figure*}
  \centering
  \begin{subfigure}{\textwidth}
    \includegraphics[width=0.48\linewidth]{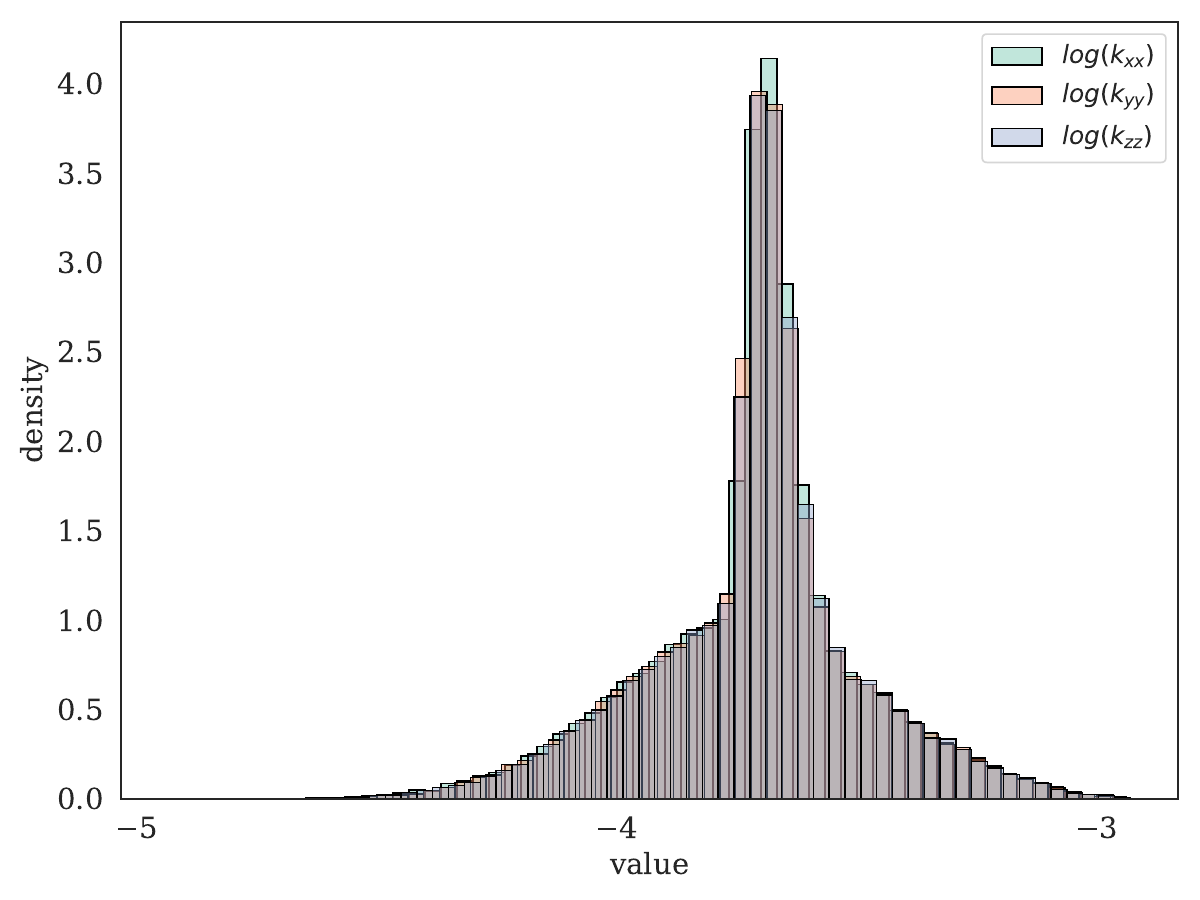}
    \includegraphics[width=0.48\linewidth]{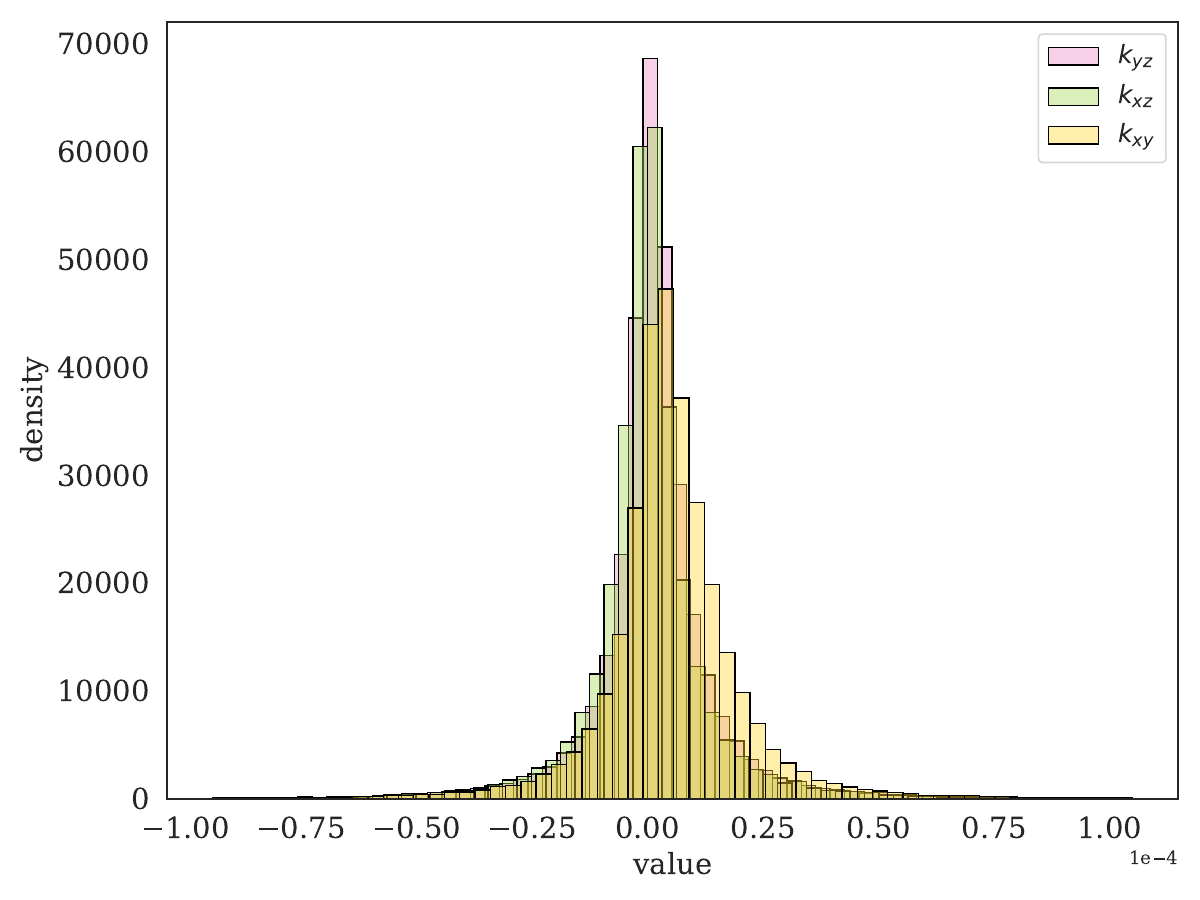}
    \caption*{Dataset $\mathcal{A}$}
  \end{subfigure}
  \centering
  \medskip

  \begin{subfigure}{\textwidth}
    \includegraphics[width=0.48\linewidth]{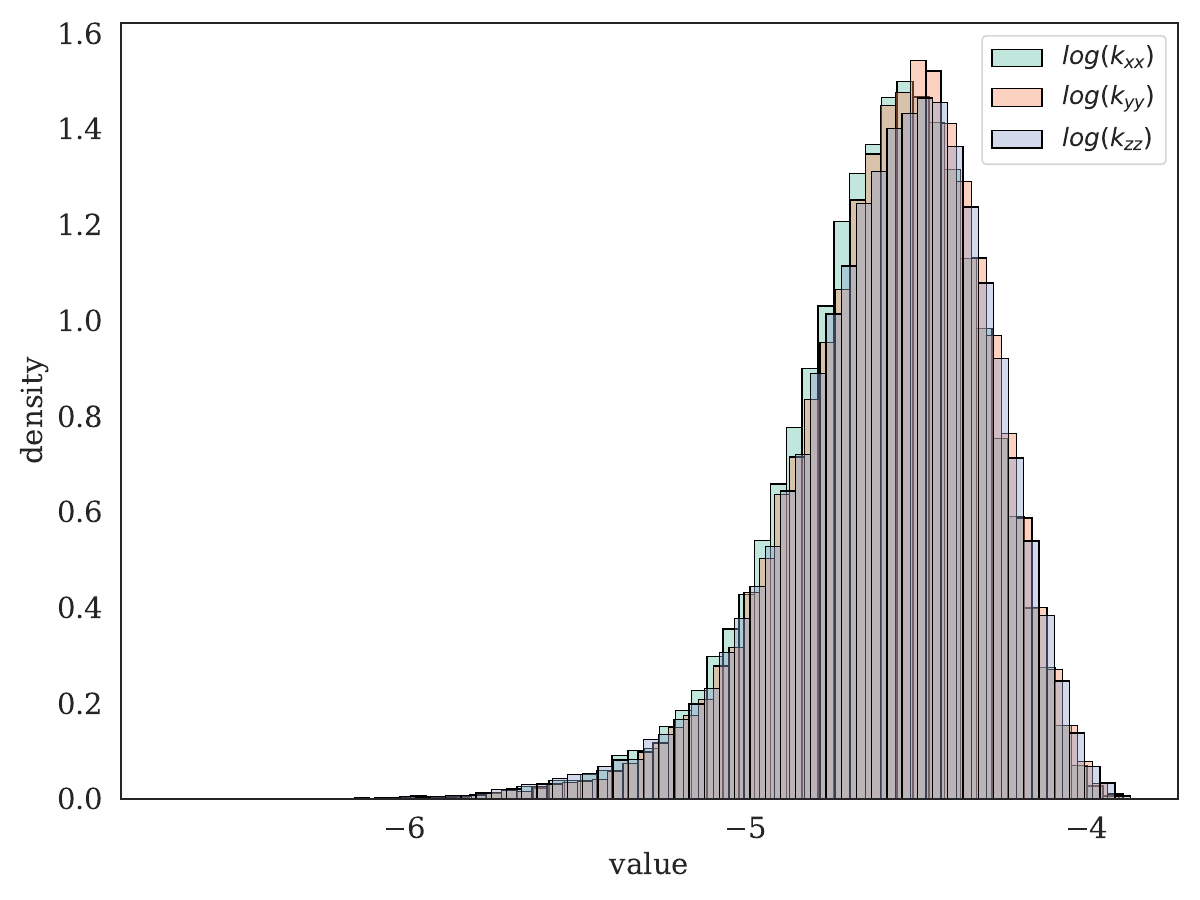}
    \includegraphics[width=0.48\linewidth]{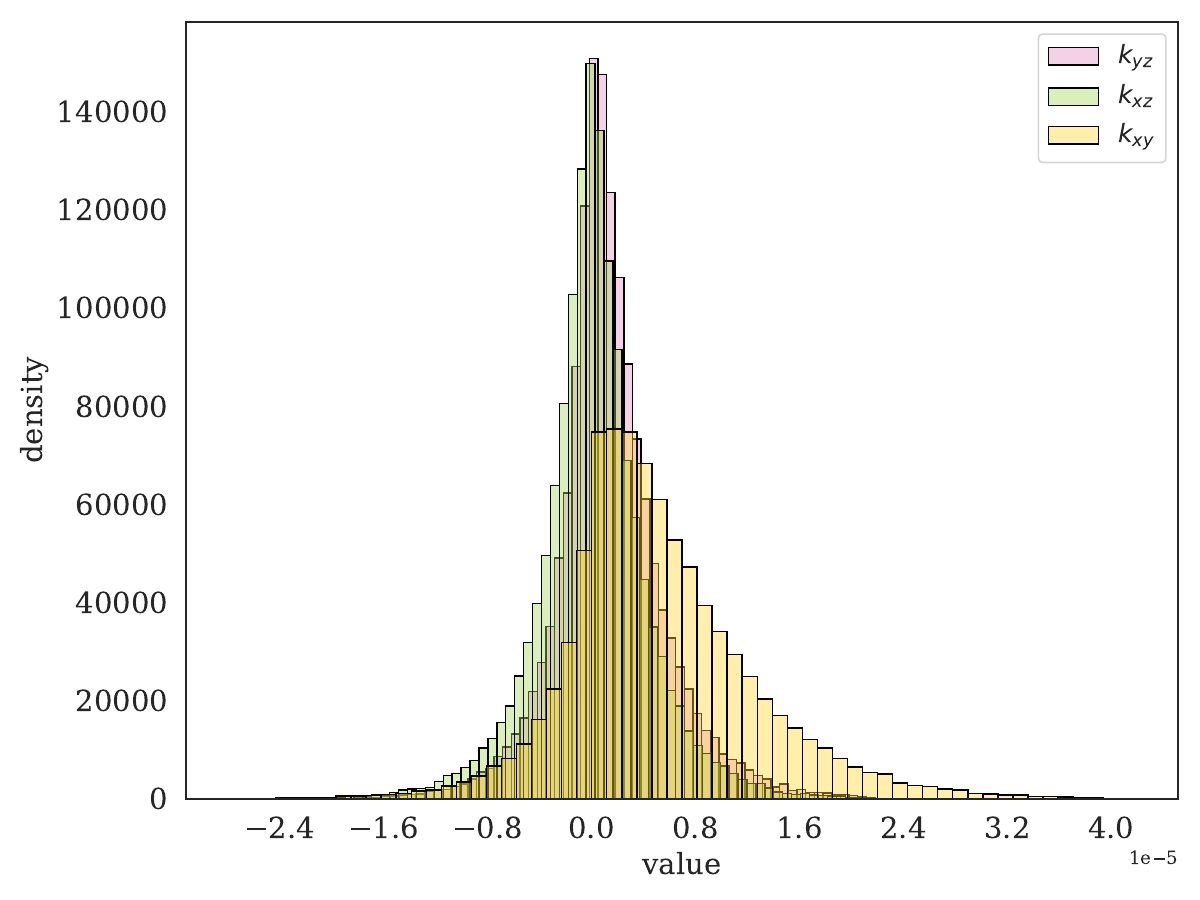}
      \caption*{Dataset $\mathcal{B}$}
  \end{subfigure}

  \medskip

  \begin{subfigure}{\textwidth}
    \includegraphics[width=0.48\linewidth]{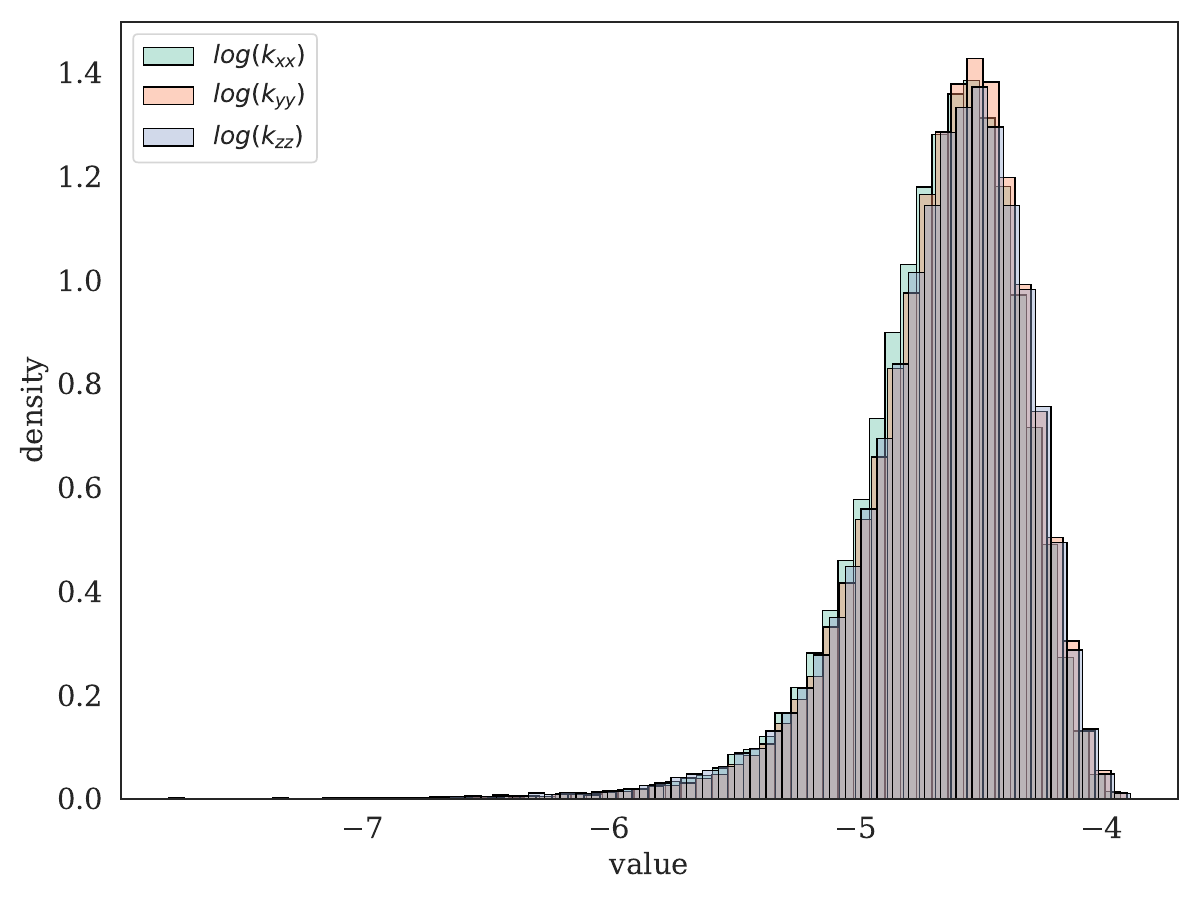}
    \includegraphics[width=0.48\linewidth]{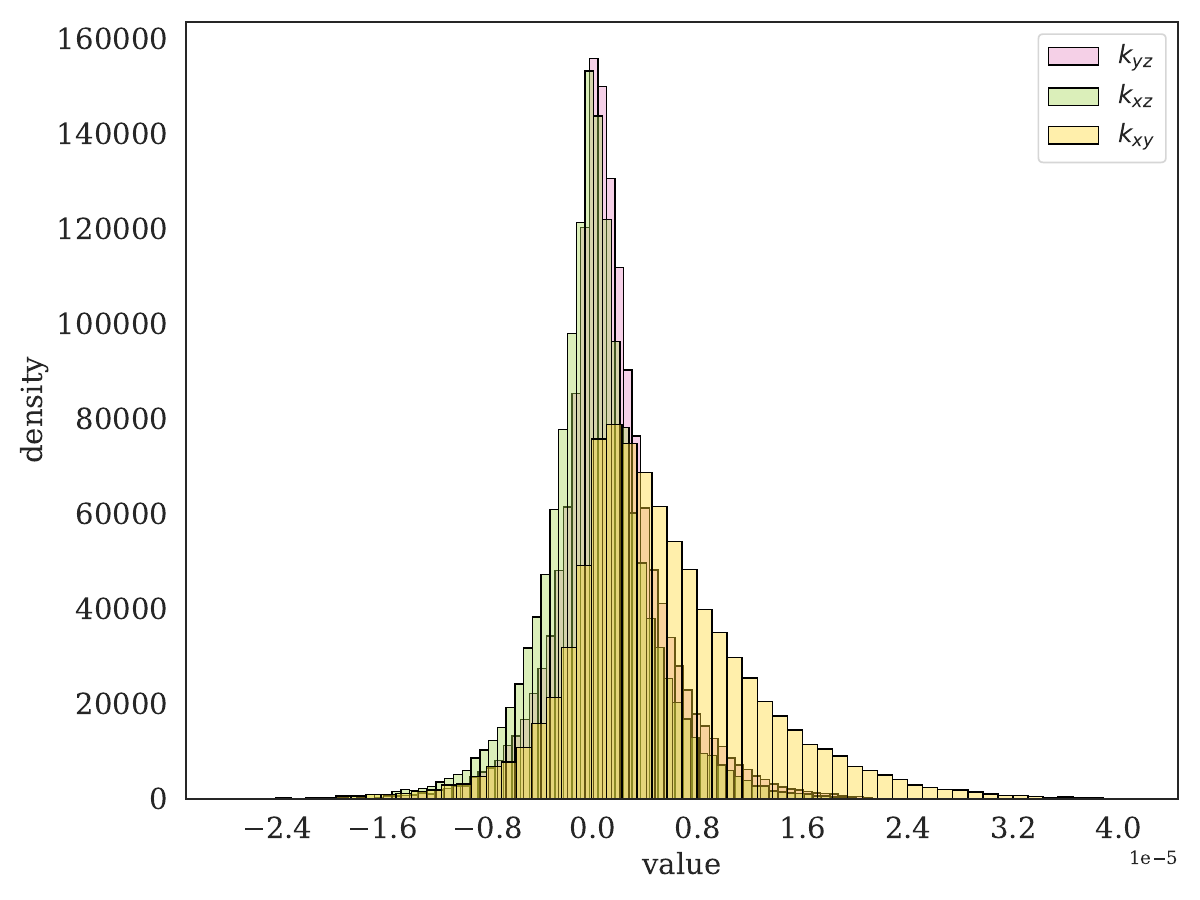}
      \caption*{Dataset $\mathcal{C}$}
  \end{subfigure}
  \caption[Distributions of $\tn K^{eq}$ components for datasets of different $K_f/K_m$]{Distributions of $\tn K^{eq}$ components for datasets of different $K_f/K_m$, Dataset $\mathcal{A}$ for $K_f/ K_m = \num{1e3}$, Dataset $\mathcal{B}$ for $K_f/ K_m = \num{1e5}$, and Dataset $\mathcal{C}$ for $K_f/ K_m = \num{1e7}$. Diagonal components are shown on a \(\log_{10}\) scale.}
  \label{fig:datasets}
\end{figure*}

The distributions of components of equivalent hydraulic conductivity tensors for datasets $\mathcal{A}$, $\mathcal{B}$, and $\mathcal{C}$ are depicted in Figure \ref{fig:datasets}. 
Focusing on the diagonal components: $k_{xx}$, $k_{yy}$, $k_{zz}$ we observe that, for Dataset $\mathcal{A}$, the equivalent tensor is primarily influenced by matrix elements characterized by $\vmu = [-4.,  -3.8, -3.9]$. 
The peak values are attributed to samples with $\lambda=0$. 
For Dataset $\mathcal{B}$, we note an increased impact of fractures. There is a heavier left tail of the distributions that accounts for samples with a low number of fractures that are not fully connected. 
As the ratio $K_f/ K_m$ increases, the left tail of the distribution becomes heavier ($K_m$ gets lower in our setting). The maximum values in $\tn K^{eq}$ do not increase, which indicates that even for $K_f/ K_m = \num{1e5}$, there are cases where the impact of bulk on $\tn K^{eq}$ is negligible. This is more pronounced for Dataset $\mathcal{C}$.
Regarding off-diagonal components: $k_{yz}$, $k_{xz}$, $k_{xy}$, the data shape is very similar for Dataset $\mathcal{B}$ and Dataset $\mathcal{C}$. While for Dataset A, the off-diagonal values are more concentrated, but of the order of magnitude higher.

The equivalent tensor distributions differ notably in both scale and shape, making it difficult to train a single surrogate model across varying $K_f/K_m$ ratios. To address this, we train separate surrogate models: Surrogate $A$ for Dataset $\mathcal{A}$, Surrogate $B$ for Dataset $\mathcal{B}$, and Surrogate $C$ for Dataset $\mathcal{C}$. For each input $\vc X_j$, the surrogate corresponding to its $K_f/K_m$ ratio is used to predict $\tn K^{eq}_{j}$.

\def\mX{\overline{\vc X}}
\subsection{Dataset preprocessing}
To facilitate surrogate training, inputs \(\vc X_j\) and outputs \(\tn K_j^{\text{eq}}\) are preprocessed.  
For each sample \(j\), the average matrix hydraulic conductivity \(\mX_j^{m}\) is computed as a scalar by averaging over all tensor components and spatial locations.
Inputs and outputs are then normalized to allow generalization across different conductivity scales for the same \(K_f/K_m\) ratio:
\[
\mathbf{X}_j = \frac{\mathbf{X}_j}{\mX_j^m}, \quad 
\mathbf{K}_j = \frac{\mathbf{K}_j}{\mX_j^m}.
\]
Second, data is standardized independently for input and output hydraulic conductivities.
This standardization ensures that all tensor components have comparable magnitudes, preventing any single component from dominating the training process. The diagonal components of the hydraulic conductivity tensor are strictly positive, often spanning several orders of magnitude. Applying a logarithmic transform, therefore, reduces their range and improves numerical stability before standardization. In contrast, the off-diagonal components take both positive and negative values, which makes the log transform inappropriate; these components are thus standardized directly.

Each component $k^j_{\alpha\beta}$ of the hydraulic conductivity tensor (where $\alpha, \beta \in \{x, y, z\}$) is standardized:
\[
\tilde{k}^j_{\alpha\beta} =
\begin{cases} 
\displaystyle \frac{\log(k^j_{\alpha\alpha}) - \hat{\mu}_{\mathcal{L}}(\log k_{\alpha\alpha})}{\hat{\sigma}_{\mathcal{L}}(\log k_{\alpha\alpha})}, & \quad \text{if } \alpha = \beta \text{ (diagonal components)} \\[15pt]
\displaystyle \frac{k^j_{\alpha\beta} - \hat{\mu}_{\mathcal{L}}(k_{\alpha\beta})}{\hat{\sigma}_{\mathcal{L}}(k_{\alpha\beta})}, & \quad \text{if } \alpha \neq \beta \text{ (off-diagonal components)}
\end{cases}
\]
where the empirical mean 
$
\hat{\mu}_{\mathcal{L}}(k_{\alpha\beta}) = \frac{1}{|\mathcal{L}|} \sum_{j \in \mathcal{L}} k^j_{\alpha\beta}
$
and empirical standard deviation \\
$\hat{\sigma}_{\mathcal{L}}(k_{\alpha\beta}) = \sqrt{\frac{1}{|\mathcal{L}| - 1} \sum_{j \in \mathcal{L}} (k^j_{\alpha\beta} - \hat{\mu}_{\mathcal{L}}(k_{\alpha\beta}))^2}
$
are estimated from the training set \( \mathcal{L} \).

\section{Surrogate architecture}\label{surrogate_arch_section}
The architecture of the surrogates is closely guided by the structure of the dataset samples. Given the nature of the input data, the surrogate combines a convolutional neural network (CNN), which acts as a feature extractor (see Section~\cite[ch.~9]{Goodfellow-et-al-2016}), with a feed-forward neural network (FNN) for the regression task (see~\cite[p.~328]{Goodfellow-et-al-2016}).
To determine the final architecture, we explored various combinations of convolutional layers, kernel sizes, and their padding settings, as well as different FNN depths and widths, several activation functions, and both max-pooling and average-pooling strategies. While some of these alternatives produced predictions with a comparable accuracy, the configuration presented below consistently offered the best balance between performance and computational efficiency.

The CNN part comprises four Conv3D layers, each integrating 3D convolution and max pooling. The convolution uses a kernel of size \((3,3,3)\), stride $1$, and padding $1$, followed by max pooling with size \((2,2,2)\) and stride $2$. Batch normalization is applied after each convolution to stabilize training.
Starting from input data of shape \(64 \times 64 \times 64 \times 6\), the CNN reduces it to a feature map of size \(4 \times 4 \times 4 \times 1296\), which is then flattened using global average pooling. The resulting vector passes through fully connected layers with $2048$, $2048$, and $1024$ neurons. So far, each layer uses ReLU activation to introduce nonlinearity. The final output layer has 6 neurons with an identity activation function. The full architecture is detailed in Table~\ref{NN_architecture}.
\begin{table}
\centering
\caption{The Architecture of the surrogate}
\label{NN_architecture}
\begin{tabular}[t]{lccc}
\hline
Layer & Type & Output size & Learnable parameters 
\\
\hline
 &  Input  & $64 \times 64 \times 64 \times 6$ & \\
1 & Conv3D & $32 \times 32 \times 32 \times 48$ & $7{,}824$\\
2 & Conv3D & $16 \times 16 \times 16 \times 144$ & $186{,}768$\\
3 & Conv3D & $8 \times 8 \times 8 \times 432$ & $1{,}680{,}048$\\
4 & Conv3D & $4 \times 4 \times 4 \times 1296$ & $15{,}117{,}840$ \\
5 & AvgPool& $1 \times 1 \times 1 \times 1296$ & \\
6 & Linear & $2048$ & $2{,}656{,}256$\\
7 & Linear & $2048$ & $4{,}196{,}352$\\
8 & Linear & $1024$ & $2{,}098{,}176$\\
9 & LinearOutput & $6$  & $6{,}150$\\
\hline
\end{tabular}
\end{table}
\FloatBarrier
Surrogates are trained for 125 epochs, with the version yielding the lowest validation loss retained for further use. 
The initial learning rate \(\alpha = 0.0025\) is adaptively reduced by 50\% if the validation loss does not improve for $10$ consecutive epochs. Training is performed using the Adam optimizer with a batch size of 64.

Mean squared error (MSE) is used as the loss function, monitored across training, validation, and test datasets. To compare models, we report the normalized root mean squared error (NRMSE), which expresses prediction error relative to target variability. For component \(\tn K^{eq}_i\), $i=1, \dots, 6$ NRMSE is defined as:
\begin{equation}
\mathrm{NRMSE}_i = \frac{ \sqrt{ \frac{1}{|\mathcal{M}|} \sum_{\mathcal{M}} \left( \tn K^{eq}_i - f_i(\mathbf{X}) \right)^2 } }{ 
\mathrm{std}_{\mathcal{M}} \big(\tn K^{eq}_i\big)},
\end{equation}
where the numerator is the root mean squared error, and the denominator is the empirical standard deviation of the target values over the test set \(\mathcal{M}\). This normalization enables meaningful comparisons of NRMSE across components and test datasets with varying output variances. For further analysis, we also compute the average NRMSE across all components:
\begin{equation}
\overline{\text{NRMSE}} = \frac{1}{6} \sum_{i=1}^6 \text{NRMSE}_i.
\end{equation}

\section{Results}\label{results_section}
\subsection{Prediction accuracy on test datasets}\label{prediction_accuracy_on_test_sets}
\begin{figure*}
  \begin{subfigure}{\textwidth}
  \centering
\includegraphics[width=0.37\linewidth]{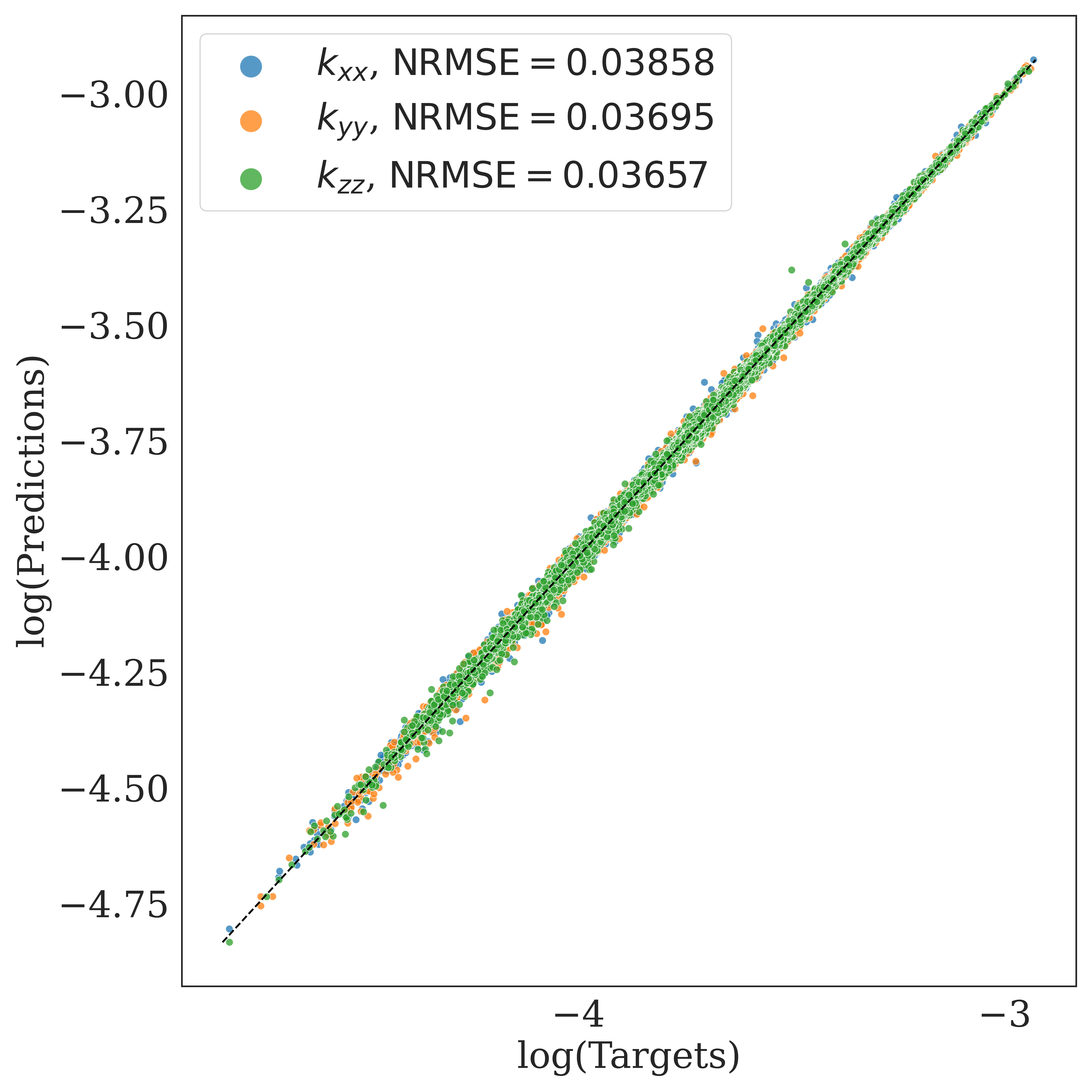}
\includegraphics[width=0.37\linewidth]{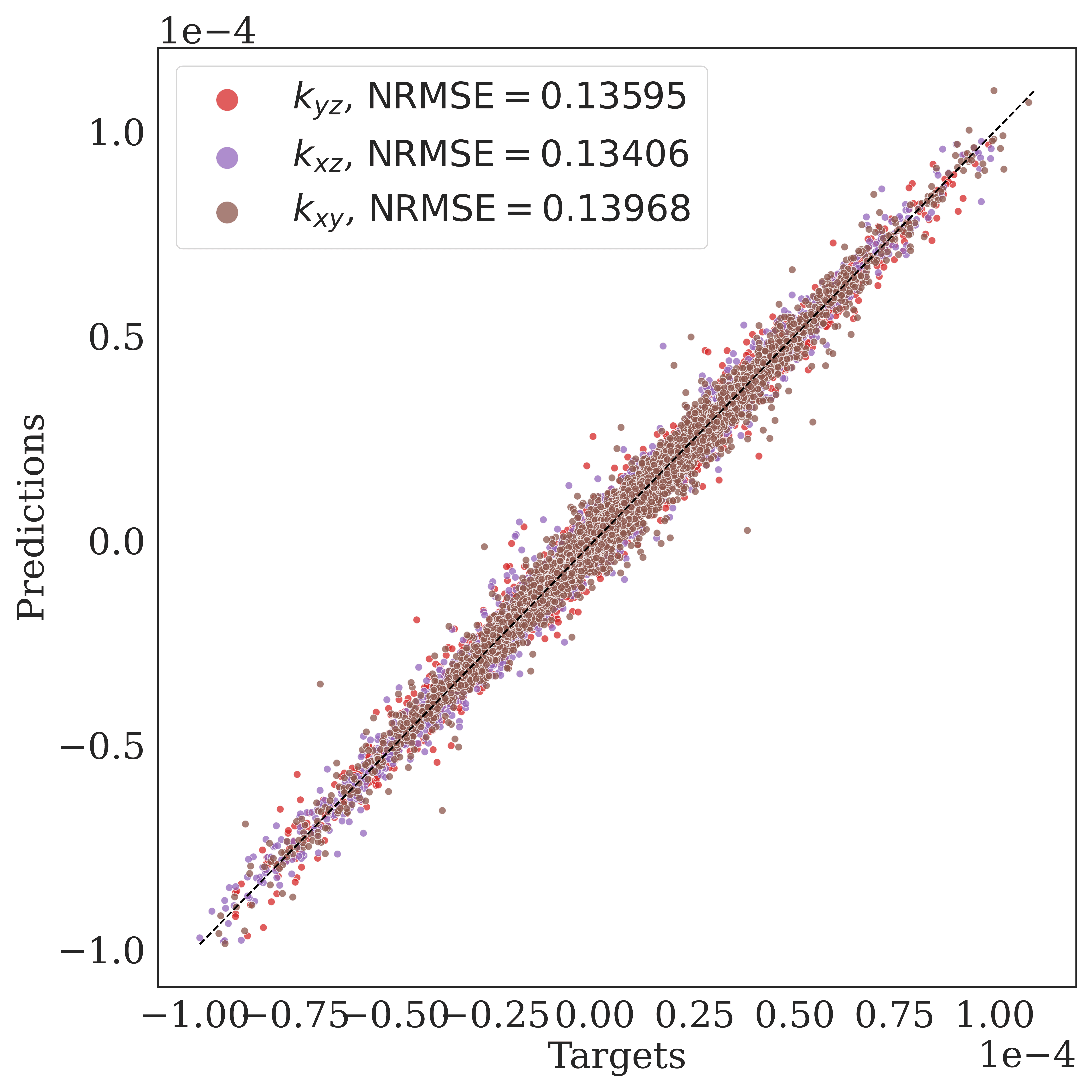}
    \caption*{Surrogate $A$}
  \end{subfigure}
  \centering
  
  \medskip

  \begin{subfigure}{\textwidth}
    \centering\includegraphics[width=0.37\linewidth]{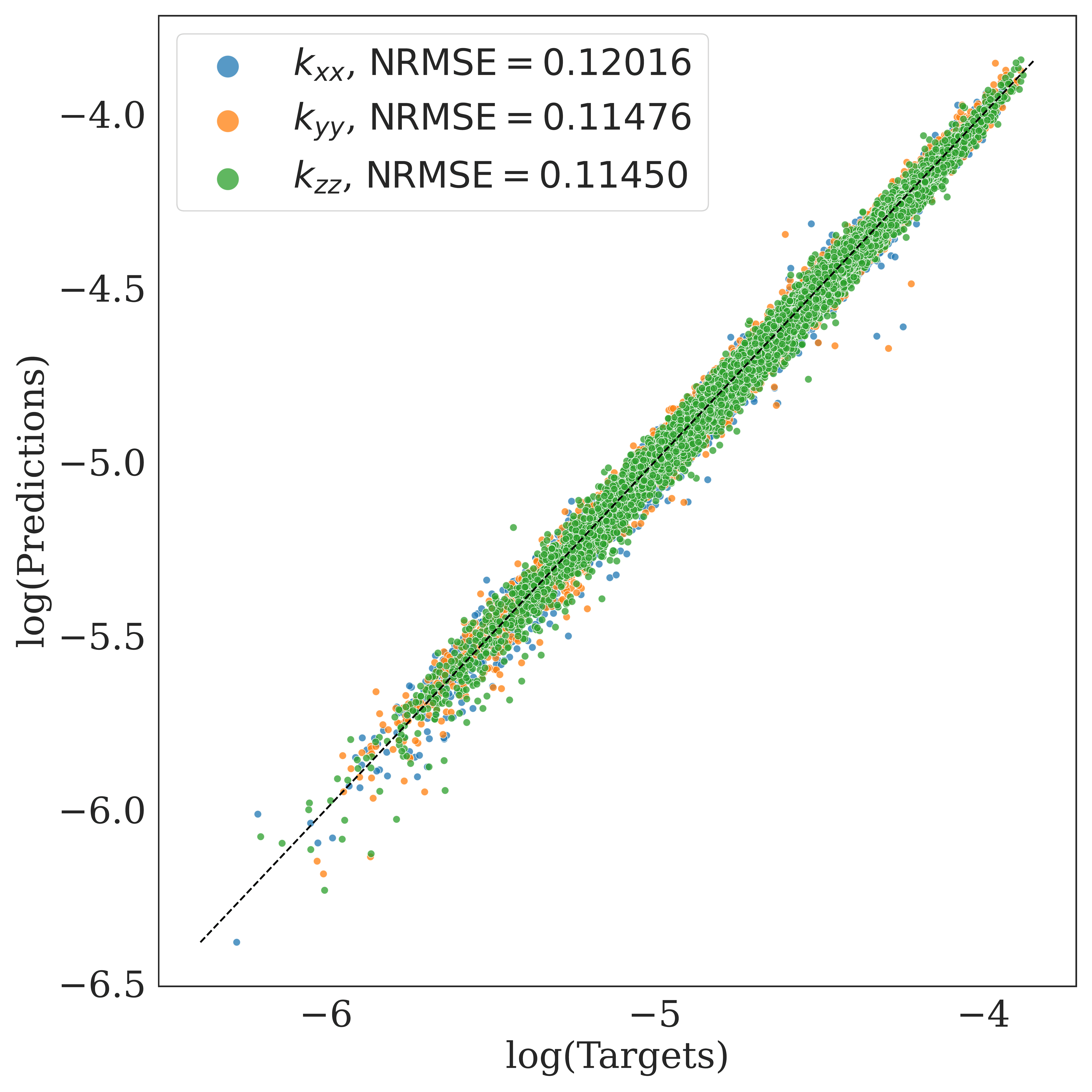}
    \includegraphics[width=0.37\linewidth]{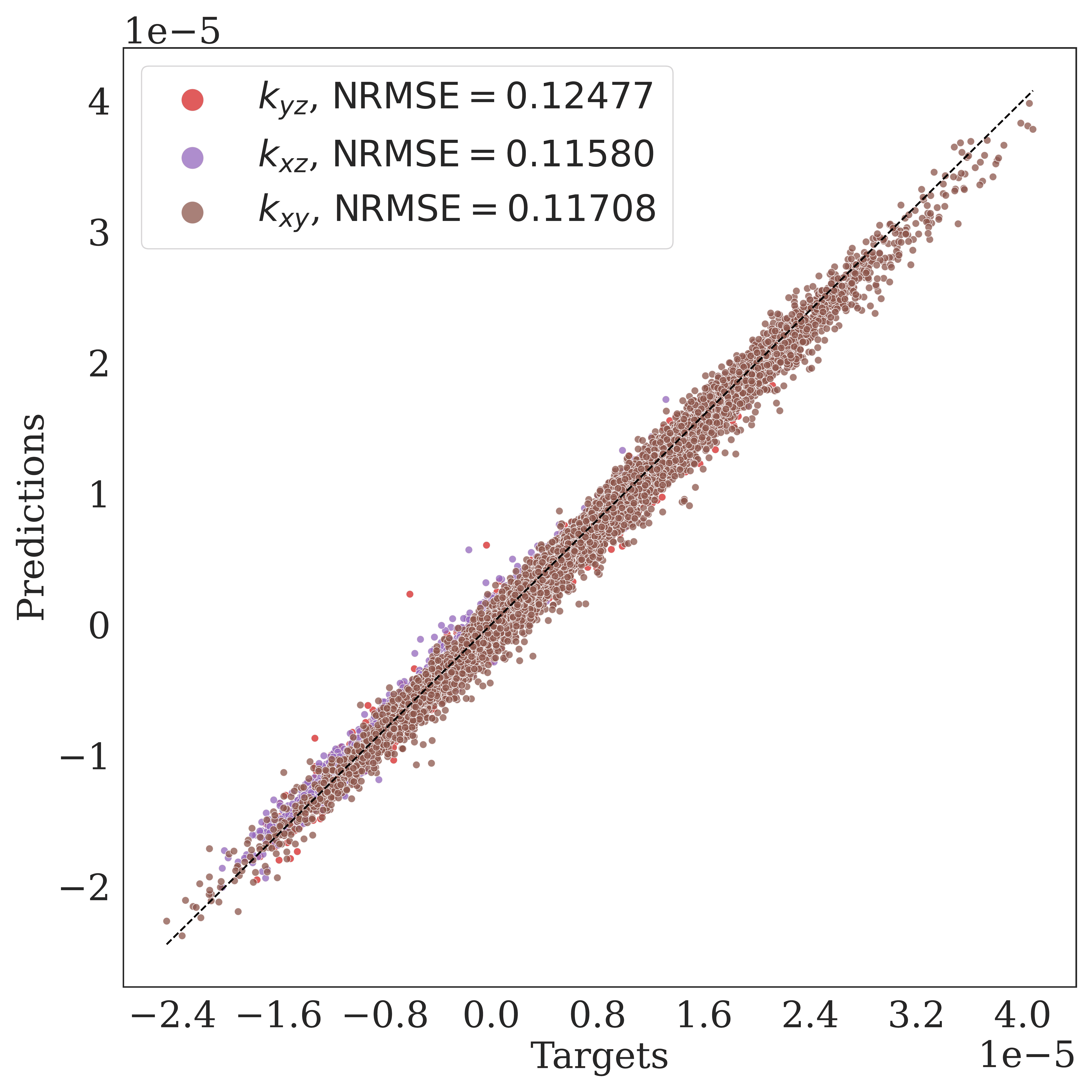}
      \caption*{Surrogate $B$}
  \end{subfigure}

  \medskip

  \begin{subfigure}{\textwidth}
\centering\includegraphics[width=0.37\linewidth]{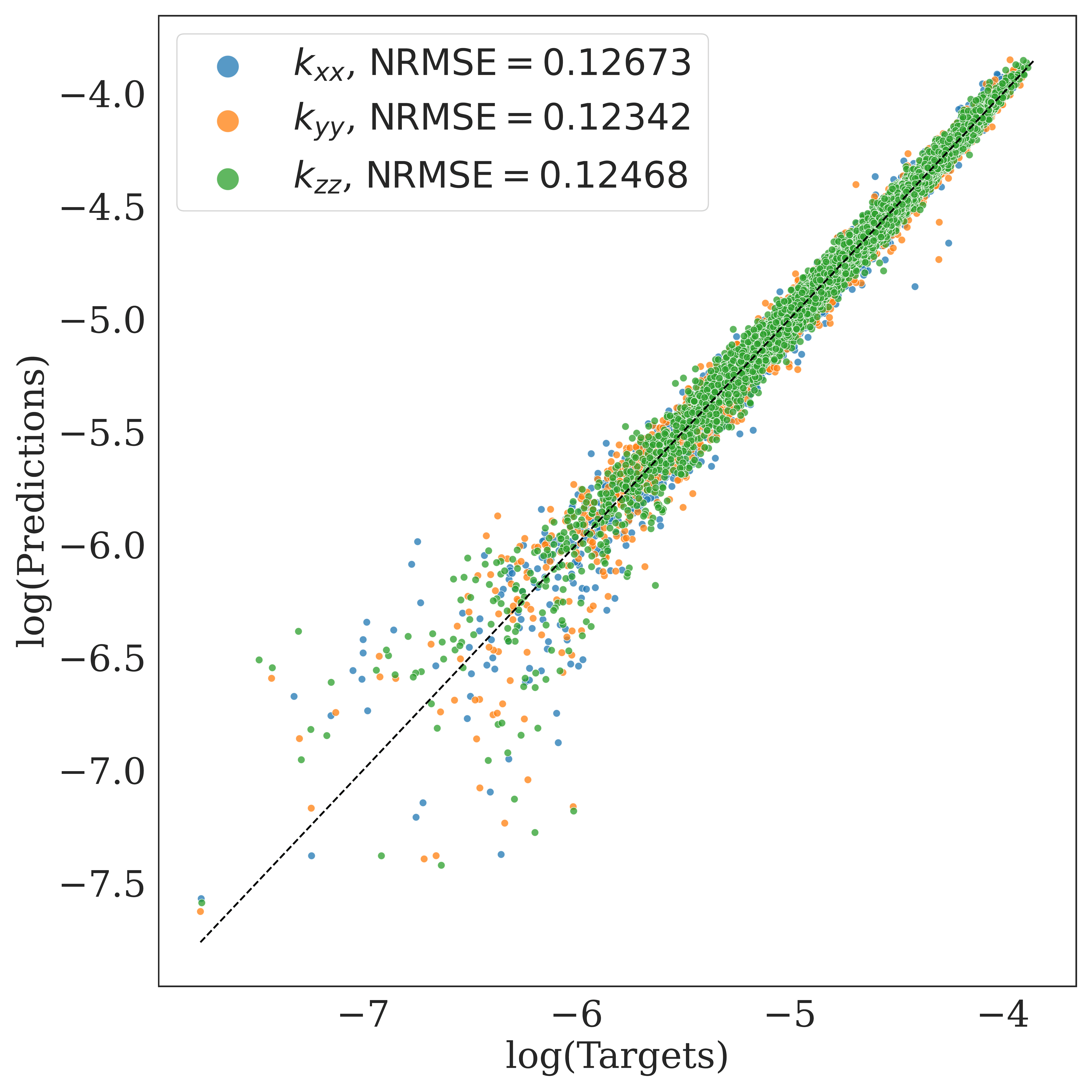}
    \includegraphics[width=0.37\linewidth]{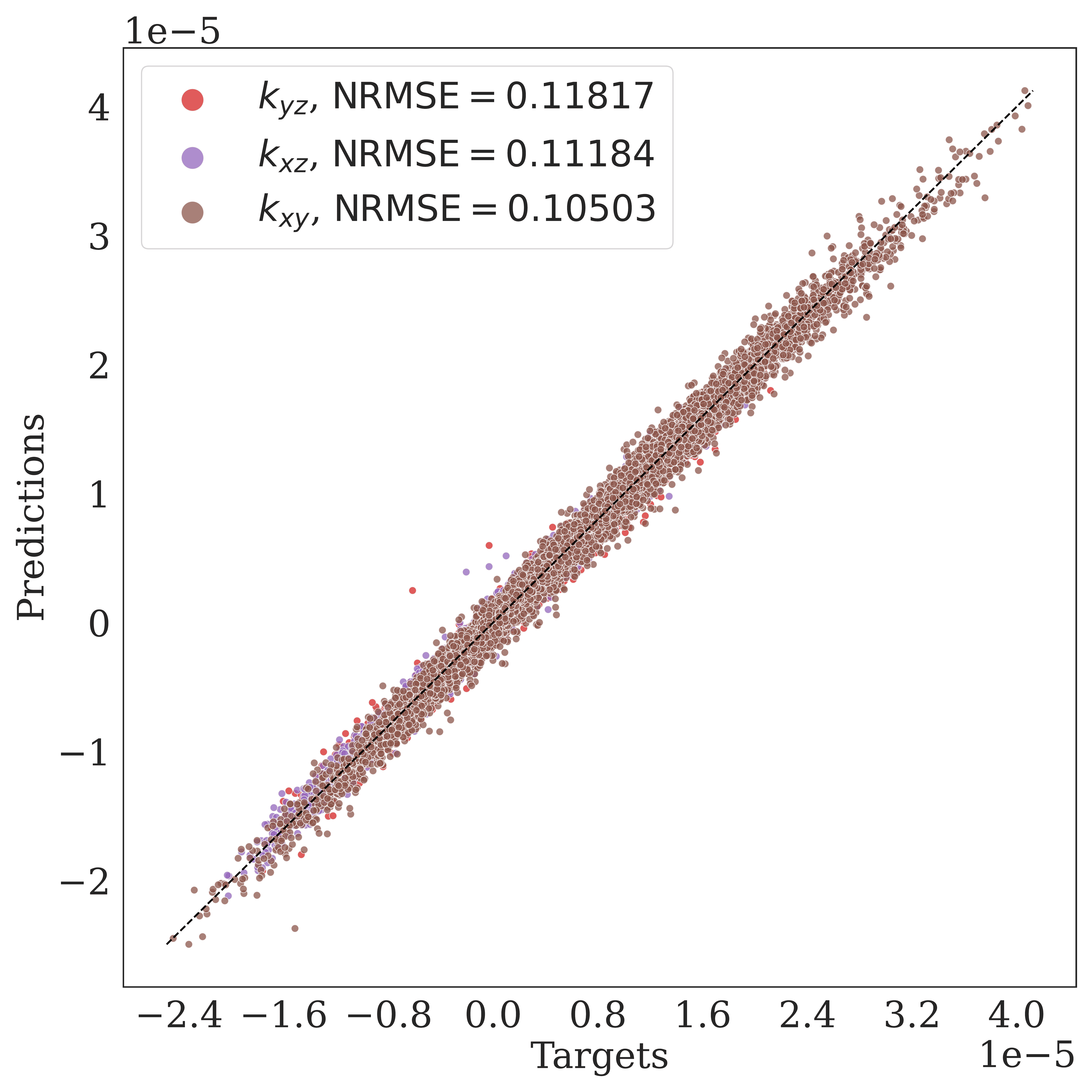}
      \caption*{Surrogate $C$}
  \end{subfigure}
    \caption[The prediction accuracy of the trained surrogates for the components of $\tn K^{eq}$ evaluated on test datasets.]{The prediction accuracy of the trained surrogates for the components of $\tn K^{eq}$ evaluated on test datasets. Diagonal components are shown on a \(\log_{10}\) scale.}
  \label{fig:fractures_diag_cond_surrogates_accuracy_mse_loss}
\end{figure*}

This section evaluates the prediction accuracy of the trained surrogates on the test datasets $\mathcal{T}_A$, $\mathcal{T}_B$, and $\mathcal{T}_C$. Figure~\ref{fig:fractures_diag_cond_surrogates_accuracy_mse_loss} displays target-versus-prediction plots and NRMSE values for each surrogate and every component of the predicted $\tn K^{eq}$.

For the diagonal components (left column), Surrogate A demonstrates the highest accuracy ($\text{NRMSE} < 0.04$), which can be attributed to the narrow distribution of these components. With training datasets of equal size, narrower distributions tend to be better represented, resulting in improved model performance. 
As the data distributions become broader and more complex (Figure~\ref{fig:datasets}), the predictive accuracy of the surrogate models declines  (Figure~\ref{fig:fractures_diag_cond_surrogates_accuracy_mse_loss}). 
This trend is especially pronounced for Surrogate C, where predictions for samples in the left tail of the distribution are considerably less accurate, likely due to their underrepresentation in the training set. This issue becomes more severe with increasing $K_f/K_m$.

In contrast, for the off-diagonal components, Surrogate~A performs the worst ($\text{NRMSE} \approx 0.14$), reflecting the broader and more complex nature of these components’ distributions. As a result, the overall prediction accuracy of Surrogate A is largely constrained by its ability to estimate off-diagonal components. Surrogate B achieves a more uniform accuracy across all components of $\tn K^{eq}$. For Surrogate C, the off-diagonal components are predicted with even slightly higher accuracy than the diagonal ones.

Although underrepresentation in certain regions of the training distribution adversely affects surrogate performance, this issue could potentially be alleviated through targeted dataset augmentation. However, we do not have a straightforward method for generating DFM homogenization blocks that correspond to a prescribed $\tn K^{eq}$.

\FloatBarrier

\subsection{Impact of correlation length of SRF on prediction accuracy}\label{prediction_srf_corr_length}
We evaluated the effect of the matrix SRF correlation length $\lambda$ on surrogate accuracy using test datasets of $3{,}500$ samples with fixed fracture density ($P_{30} = 0.0015$) and $\lambda \in \{0, 5, 10, 15, 25, 50, 75, 100, 250, 500, 1000\}$. Surrogates were trained on data with $\lambda \in \{0, 10, 25\}$ equally represented.

\begin{figure}
 \centering
 \includegraphics[width=0.8\textwidth]{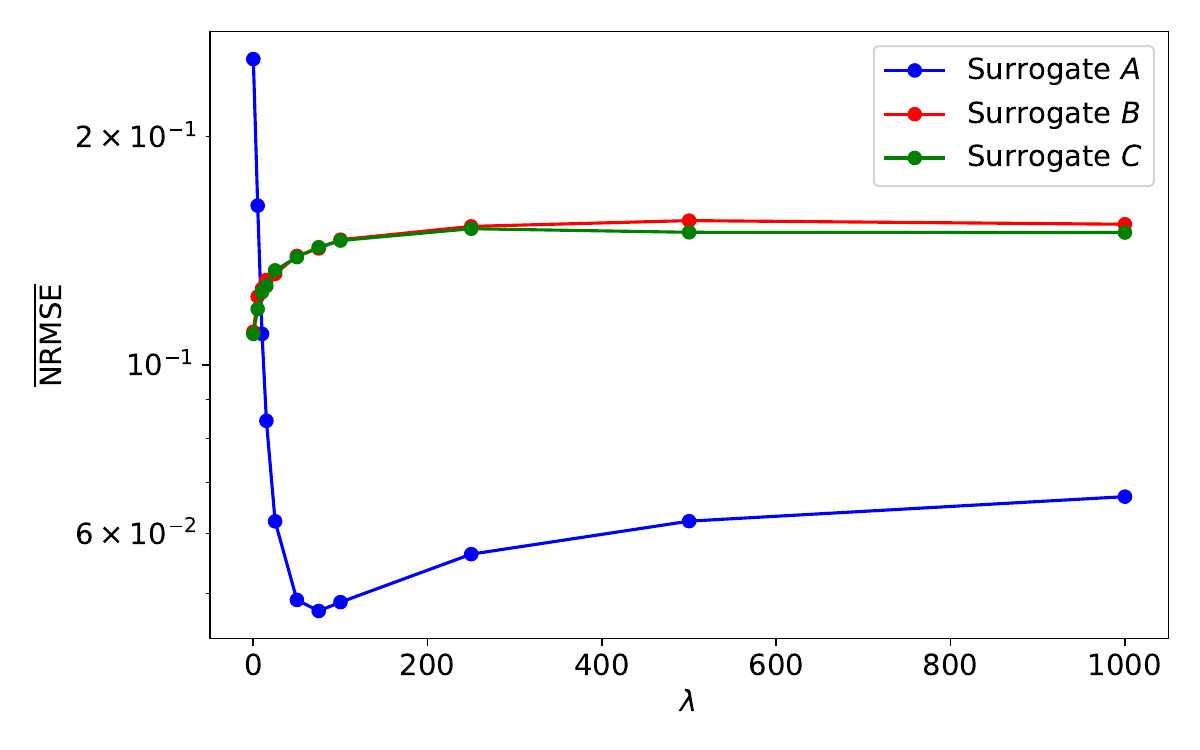}
 \caption{Effect of the SRF correlation length $\lambda$ on the prediction accuracy of the surrogate model.}
 \label{fig:fractures_diag_cond_nrmse_corr_len}
\end{figure}

As shown in Figure~\ref{fig:fractures_diag_cond_nrmse_corr_len}, the prediction accuracy ($\overline{\text{NRMSE}}$) of Surrogate A improves with increasing $\lambda$ up to 75, consistent with reduced SRF variability at longer correlation lengths and limited fracture influence, which simplifies the learning task. Accuracy slightly declines beyond $\lambda=75$, likely due to limited training data.
In contrast, Surrogates B and C show decreasing accuracy with increasing $\lambda$. Trained on high fracture–matrix conductivity ratios ($K_f/K_m = \num{1e5}$ and $\num{1e7}$), they struggle to distinguish samples with similar DFNs but varying SRFs, particularly at large $\lambda$ values underrepresented in training datasets. Both perform best for uncorrelated fields ($\lambda=0$).

All surrogates generalize well to intermediate $\lambda = \{5, 15\}$, indicating that training on $\lambda = \{0, 10, 25\}$ sufficiently captures the range of relevant SRF correlation lengths within the $15 \times 15 \times 15$ homogenization block. Similar trends were observed in our 2D study \cite{Spetlik2024}. 
Henceforth, $\lambda=10$ is used for testing, as it yields consistent accuracy across all surrogates.

\subsection{Impact of number of fractures}
We further analyzed the prediction accuracy of the surrogates for test datasets of a different number of fractures. Originally, surrogates were trained on the dataset with samples of $P_{30}= 0.0010$ and $P_{30}=0.0025$ equally represented. 
Test datasets of $3{,}500$ samples, $\lambda=10$ and $P_{30} \in \{0.0004, 0.0006, 0.0008, 0.0012, 0.0015, 0.0020, 0.0025,  0.0030, 0.0050,\\ 0.0075, 0.0100, 0.0150, 0.0200\}$ were formed.

\begin{figure}
 \centering
 \includegraphics[width=0.97\textwidth]{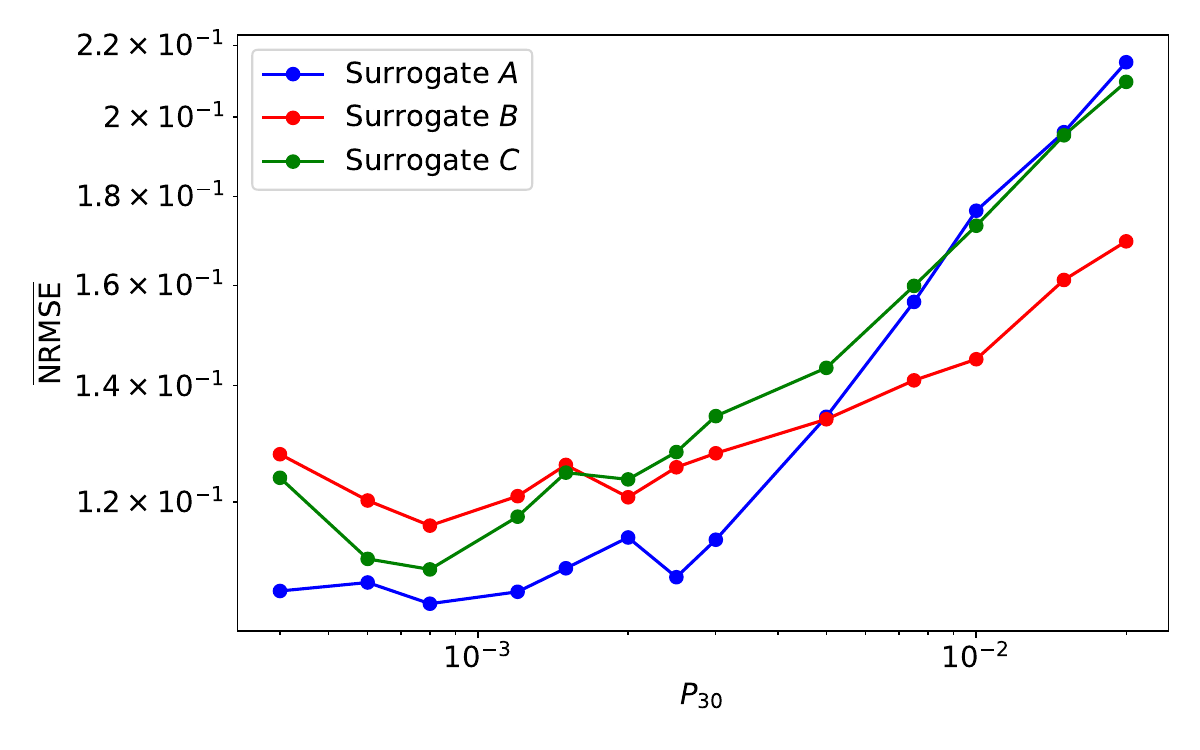}
 \caption{Impact of number of fractures ($P_{30}$) on surrogate prediction accuracy.}
 \label{fig:fractures_diag_cond_nrmse_P_30}
\end{figure}

Figure~\ref{fig:fractures_diag_cond_nrmse_P_30} shows the development of $\overline{\text{NRMSE}}$ depending on $P_{30}$. All three surrogates achieve their best prediction accuracy at $P_{30} = 0.0008$.
For Surrogate A, $\overline{\text{NRMSE}}$ remains nearly constant for $P_{30} \in \{0.0004, 0.0006, 0.0008, 0.0012\}$, followed by a sharp increase, except at $P_{30}=0.0025$, which was explicitly included for training. Even for $P_{30}=0.0200$, Surrogate A maintains a reasonable accuracy with $\overline{\text{NRMSE}}=0.2151$.
Surrogates B and C show a different trend: their accuracy slightly improves from $P_{30} = 0.0004$ to $P_{30} = 0.0008$, followed by an increase that drops at $P_{30}=0.0020$. Surrogate A is the most sensitive to $P_{30}$ values outside its training range. Surrogate B demonstrates greater robustness to changes in $P_{30}$.
The sharper increase in $\overline{\text{NRMSE}}$ for $P_{30} > 0.0020$ for Surrogate C compared to Surrogate B is driven by errors in predicting the diagonal components.

Overall, the surrogates achieve prediction accuracy with $\overline{\text{NRMSE}} < 0.22$ across all investigated $P_{30}$ values. This level of accuracy is likely sufficient for a wide variety of applications. If higher accuracy is desired, the training dataset can be rebalanced by incorporating samples covering a broader range of $P_{30}$ values.

\subsection{Prediction accuracy on datasets of different DFN settings}
To evaluate the prediction accuracy across varying DFN parameter settings, we generated four additional test datasets, each comprising $3{,}500$ samples with $\lambda=10$ and $P_{30}=0.0015$.
Table~\ref{tab:test_DFN_configurations} summarizes the DFN configurations. DFN 1 and DFN 2 follow the hydrological DFN model by \cite{SKB-R-09-20}[Table 2-3], developed for groundwater flow simulations in granite bedrock at the Forsmark nuclear power plant site in Sweden. 
DFN 3 and DFN 4, adopted from site descriptive modeling of the Laxemar area (\cite{SKB-R-08-78}), also represent granite bedrock. DFN 3~\cite[Table 10-31]{SKB-R-08-78} was used for upscaling studies,
while DFN 4~\cite[Table 10-16]{SKB-R-08-78} was used in a model for estimating open fracture intensity.
\begin{table}
\centering
\caption{Test DFN configurations}
\label{tab:test_DFN_configurations}
\begin{tabular}[t]{llcccc}
\hline
 DFN & Fracture  & Orientation & Size model & $P_{32}$ (mean fracture  & $P_{30}$\\
& set name  & (trend, plunge), & power-law  &  surface area)  & (fracture intensity) \\
 & & concentration & $k_r$ & ($r_0 = 1$, $564$ m) & ($r_0 = 1$, $564$ m) \\
\hline
\hline
& NS &  (292, 1) 17.8  & 2.50 & 0.073 & $0.0152$\\
 & NE & (326, 2) 14.3  &  2.70  & 0.319 & $0.0837$\\
DFN 1 & NW & (60, 6)  12.9  & 3.10 & 0.107 & $0.0380$ \\
& EW & (15, 2)  14.0  &  3.10 & 0.088  & $ 0.0313$\\
& HZ & (5, 86)  15.2  &  2.38 & 0.543  & $0.0953$\\
\hline
& NS &  (292, 1) 17.8  &  2.50 & 0.142 & $0.0296$\\
 & NE & (326, 2) 14.3  &  2.70  & 0.345 & $0.0905$\\
DFN 2 & NW & (60, 6) 12.9  &  3.10 & 0.133 & $ 0.0472$ \\
& EW & (15, 2) 14.0  &  3.10 & 0.081  & $0.0288$\\
& HZ & (5, 86) 15.2  & 2.38 & 0.316  & $0.0554$\\
\hline
& NNE &  (283, 6.5) 13  & 2.50 & 0.230 & $ 0.0480$ \\
 & ENE & (332, 0.5) 21  &  2.45 & 0.180 & $0.0351$ \\
DFN 3 & WNW & (201, 5) 12  &  2.30 & 0.190 & $ 0.0291$  \\
& NNW & (246, 10) 12  &  2.30 & 0.050  & $0.0077$ \\
& Sub-H & (334, 87) 10  & 2.20 & 0.350  & $0.0443$ \\
\hline
& ENE &  (340.3, 1.2) 15  &  2.60 & 0.390 & $0.0921$ \\
& WNW & (208.9, 2.2) 10.9  &  2.30  & 0.550 & $0.0843$ \\
DFN 4  & NS & (272.8, 12) 11.5  &  2.70 & 0.320 & $0.0840$ \\
& SubH & (277.1, 84.3) 11.1  &  2.65 & 0.290 & $0.0723$ \\
\end{tabular}
\end{table}

\begin{figure*}
  \centering
  \begin{subfigure}{\textwidth}
    \centering\includegraphics[width=0.37\linewidth]{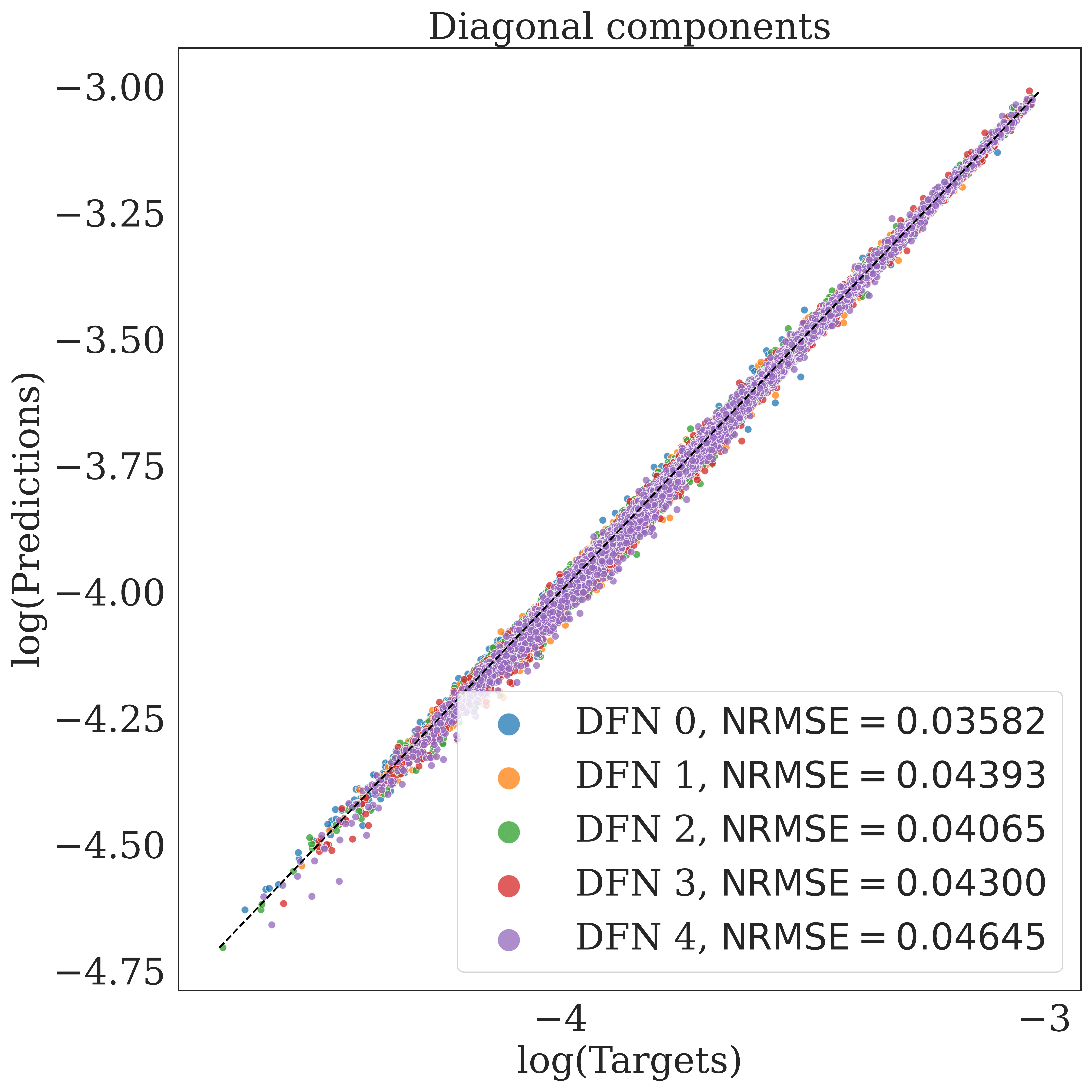}
    \includegraphics[width=0.37\linewidth]{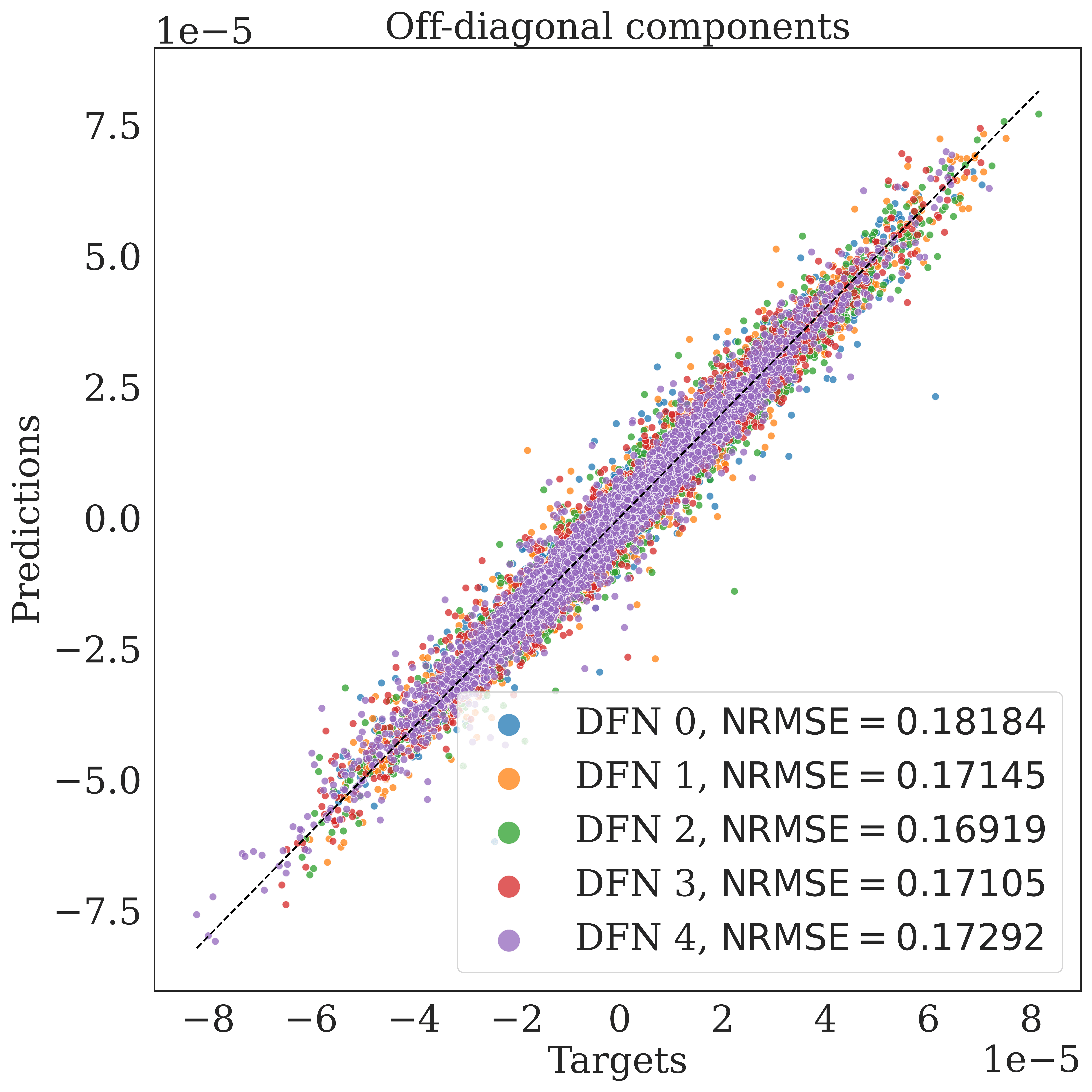}
    \caption*{Surrogate $A$}
  \end{subfigure}
  \centering
  \medskip

  \begin{subfigure}{\textwidth}
    \centering\includegraphics[width=0.37\linewidth]{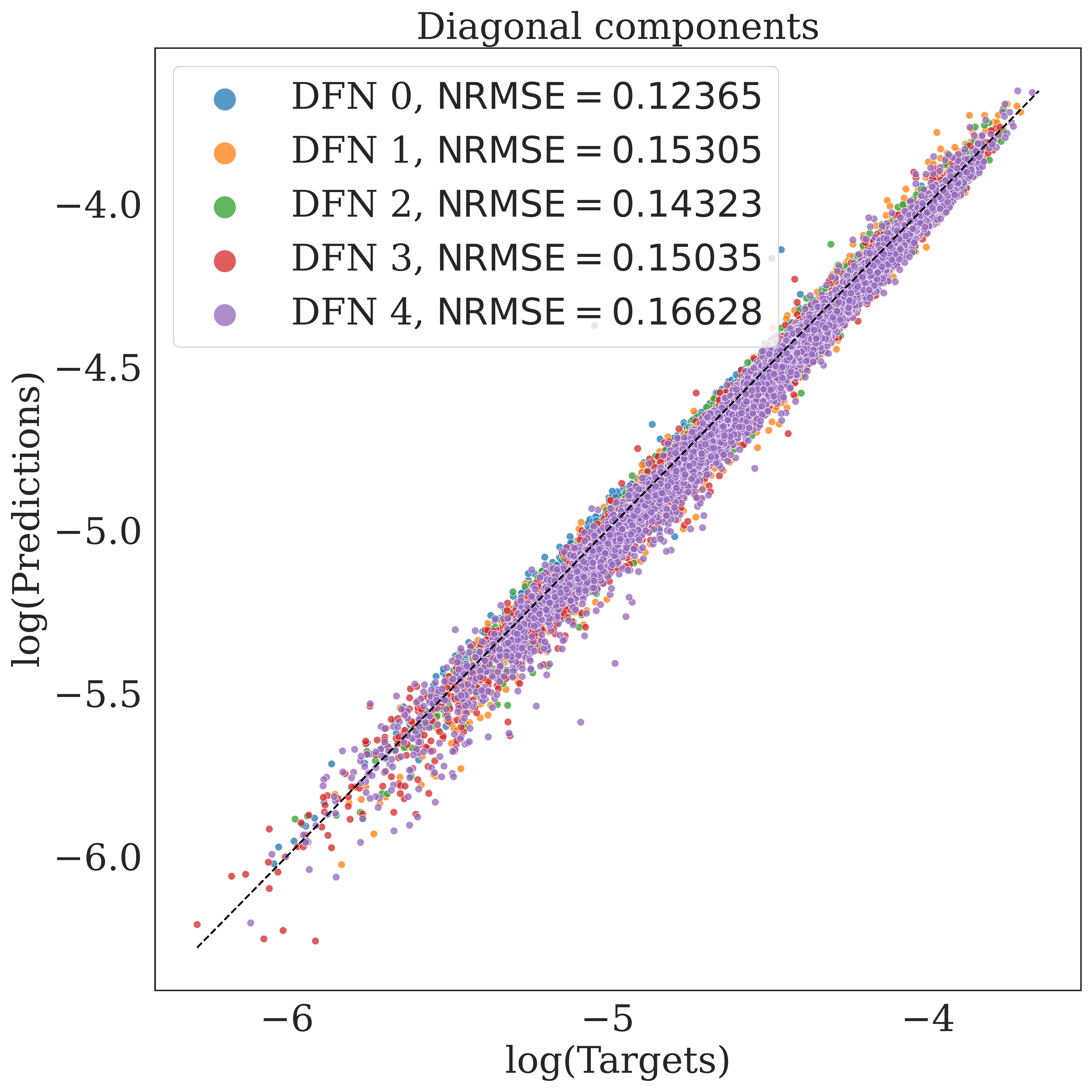}
    \includegraphics[width=0.37\linewidth]{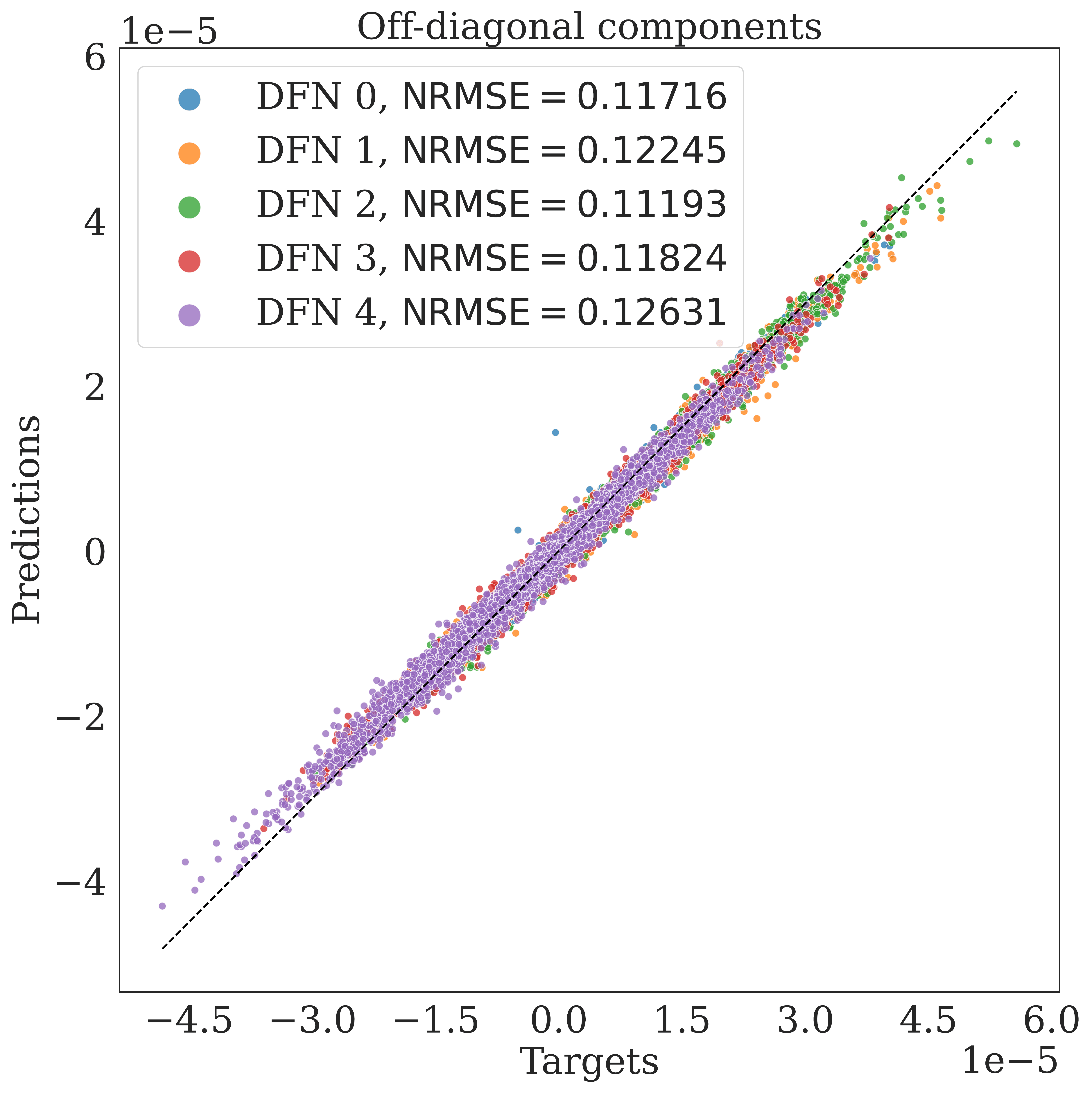}
    \caption*{Surrogate $B$}
  \end{subfigure}

  \medskip

  \begin{subfigure}{\textwidth}
    \centering\includegraphics[width=0.37\linewidth]{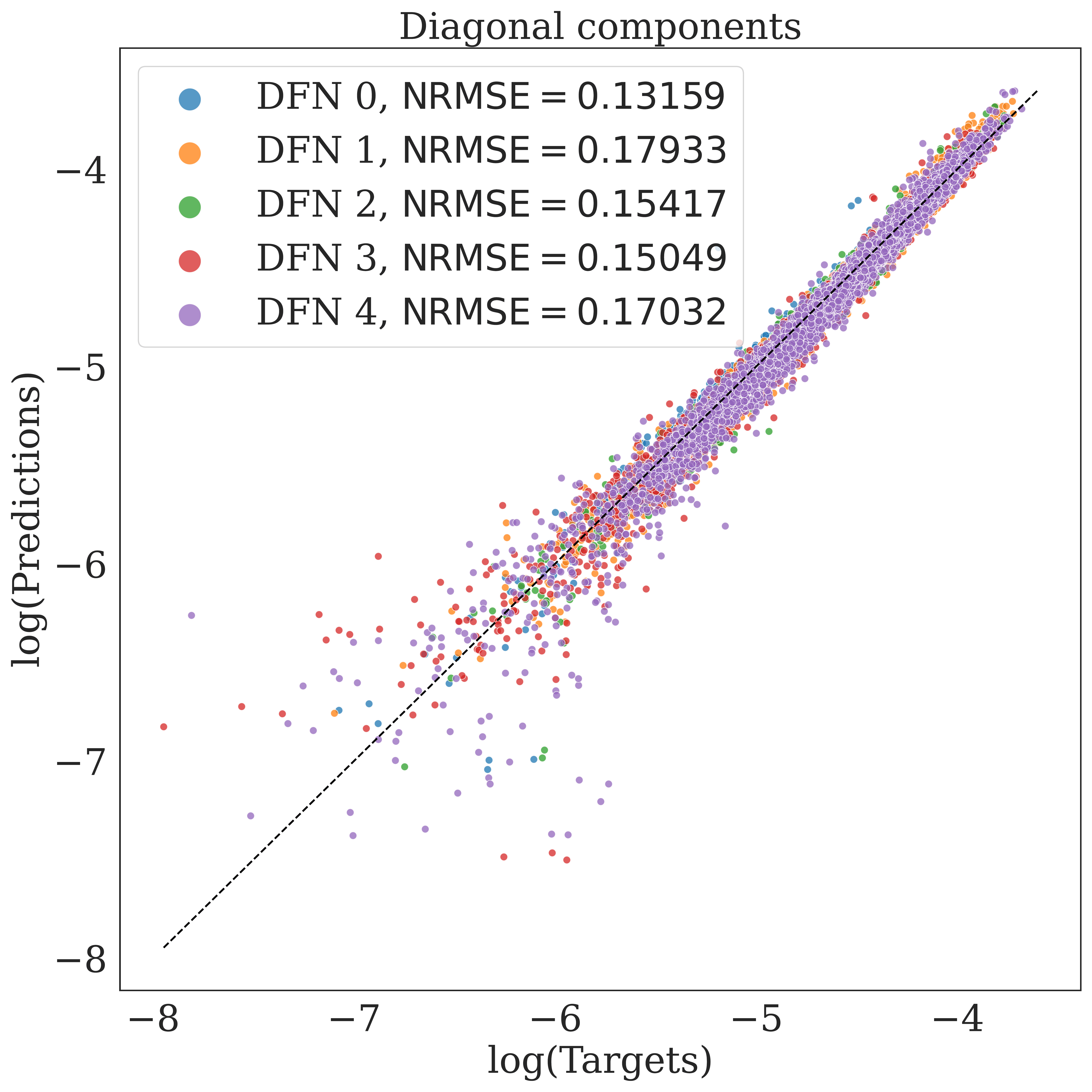}
    \includegraphics[width=0.37\linewidth]{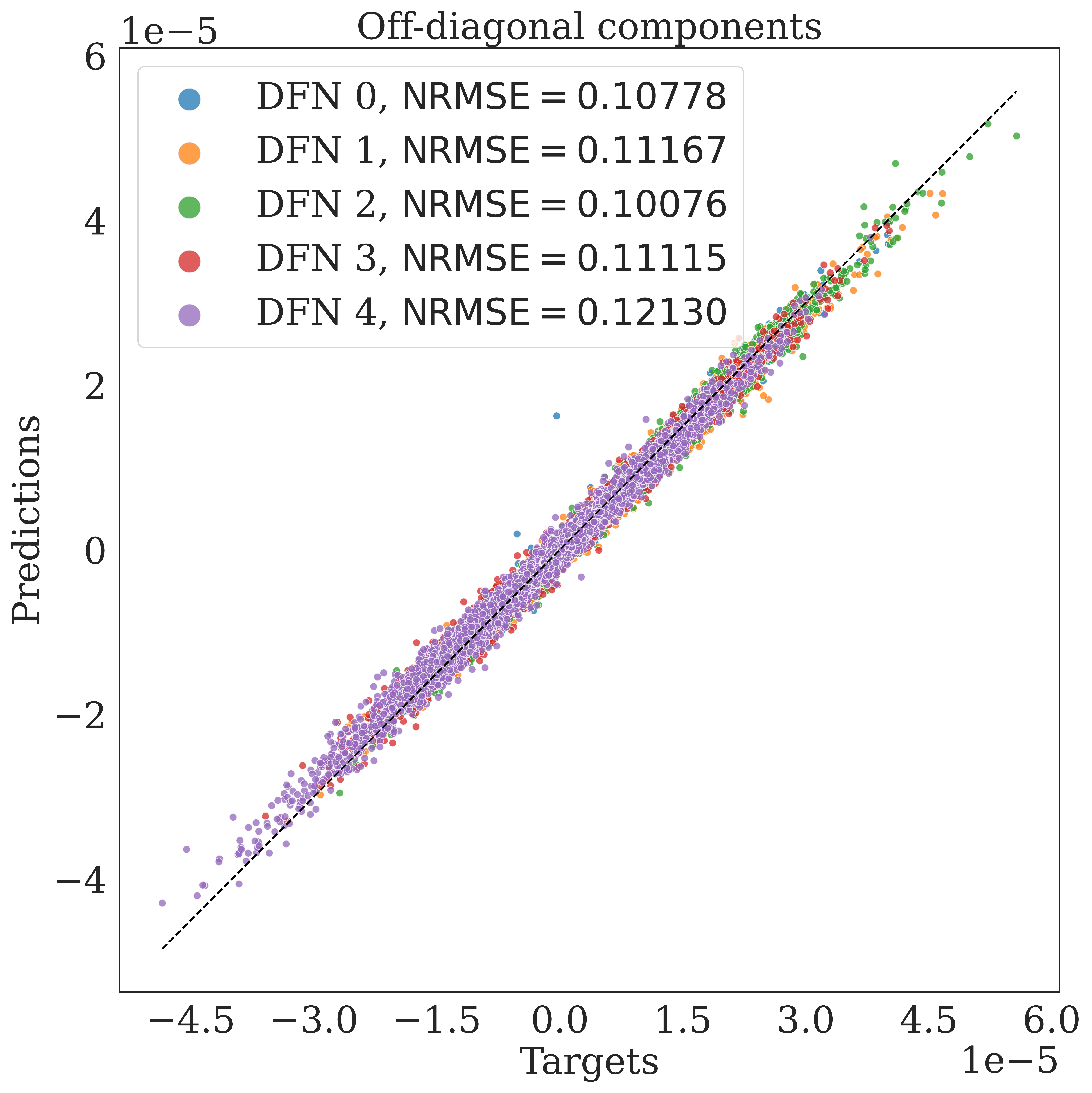}
    \caption*{Surrogate $C$}
  \end{subfigure}
    \caption[The prediction accuracy of the trained surrogates for diagonal and off-diagonal components of $\mathbf{K}^{eq}$ evaluated on test datasets with different DFNs. Diagonal isotropic $\tn K_{fr}$]{The prediction accuracy of the trained surrogates for diagonal and off-diagonal components of $\mathbf{K}^{eq}$ evaluated on test datasets with different DFNs. Diagonal components are shown on a \(\log_{10}\) scale.}
  \label{fig:fractures_diag_cond_r2_loss_DFN_params}
\end{figure*}

Figure~\ref{fig:fractures_diag_cond_r2_loss_DFN_params} compares prediction accuracy ($\text{NRMSE}$) across different DFN configurations (DFN 1–4), alongside the training setup (DFN 0). Results are shown separately for the diagonal components ($k_{xx}, k_{yy}, k_{zz}$) and the off-diagonal components ($k_{yz}, k_{xz}, k_{xy}$) of $\tn K^{eq}$, as predicted by Surrogates A, B, and C.
Off-diagonal predictions exhibit consistent accuracy across all DFN configurations for each surrogate, with minor variations likely due to dataset size and potential outliers. 
In contrast, diagonal predictions vary more significantly. Surrogates show a slight increase in $\text{NRMSE}$ for DFN 1–4. This is attributed to rare but significant fracture effects that form distribution tails, which amplify error metrics.

Despite the differences among DFN configurations, all surrogates maintain robust performance. 
This indicates that fractures within the homogenization block do not exhibit patterns that can impair neural networks' generalization capability.

\subsection{Computational cost reduction}
Surrogates are primarily employed to accelerate numerical homogenization. To quantify this benefit, we compare the computational cost of standard numerical homogenization ($C_H$ - CPU time) with that of surrogate inference ($C_S$ - GPU time), excluding training time.
We examine cases with varying domain sizes $\Omega$, corresponding to different numbers of homogenization blocks placed according to the procedure described in Section~\ref{num_hom_multiscale}. Numerical homogenization was performed on an Intel Xeon Silver 4114 CPU (2.2 GHz, 45 GB RAM), while surrogate GPU inference used an NVIDIA Tesla T4 (16 GB). Preprocessing steps such as SRF and fracture generation are excluded from the comparison, as they are identical across cases.

\begin{figure}
  \centering
    \centering
    \includegraphics[width=0.8\textwidth]{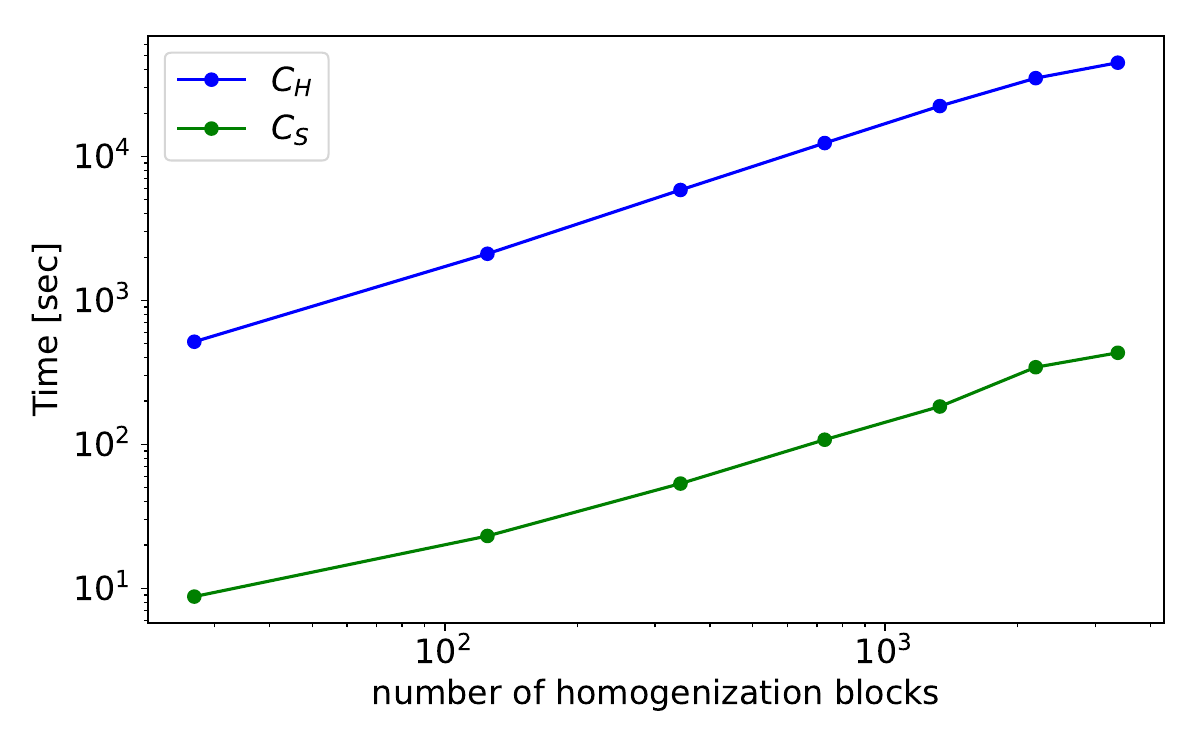}
    \caption{Comparison of computational time for numerical homogenization \(C_{H}\) (CPU run) and its surrogate counterpart \(C^{S}\) (GPU used for inference). Data shown on a \(\log_{10}\) scale.}
    \label{fig:comp_cost_surrogate}
\end{figure}

Figure~\ref{fig:comp_cost_surrogate} presents the computational times for both approaches. 
As expected, the cost increases with domain size, with standard numerical homogenization being the most computationally expensive. The cost $C_{H}$ includes mesh and SRF generation ($\approx 60\%$) and simulation runs for each block ($\approx 30\%$).
In contrast, the cost $C_S$ is dominated by voxelization (about $58\%$), with the remainder attributed to data handling; neural network inference time is negligible. The entire domain is voxelized once and split into blocks, which explains the use of 45 GB RAM. If memory becomes a limiting factor, voxelization can be performed in smaller chunks. 
We gained a speedup of $C_{H}/C_{S} > 100$ for all investigated cases. Inference on CPU reduces $C_{H}/C^{CPU}_{S} > 16$.
\FloatBarrier

\subsection{Upscaled equivalent hydraulic conductivity tensors in use}
Although the surrogates demonstrate high predictive accuracy ($\text{NRMSE} < 0.22$, corresponding to a coefficient of determination $R^2 > 0.95$), the precision required for practical applications remains uncertain. 
To investigate this, we examine how surrogates' predicted equivalent tensors for $\tn{K}_H$ affect the Constraint and Anisotropy problems described in Section \ref{sec:DFM}. 
Following the procedure outlined in Section \ref{num_hom_multiscale}, we interpolated the grid of homogenized/predicted $\tn{K}_H$ values onto the barycenters of the bulk elements in the unstructured DFM mesh. In each case, the outputs of the macro-scale DFM models computed using purely numerical homogenization were used as target values.

We consider DFM samples characterized by $P_{30} = 0.0015$, DFN 0 configuration, $\lambda=10$, a fine mesh step size of $h=5$, and a coarse mesh step size $H=10$. Trained Surrogate A, Surrogate B, and Surrogate C are employed to predict $\tn K^{eq}$, which are subsequently assembled into $\tn K_H$.  
Two domain sizes are analyzed: $\Omega_{\text{UP}} = (60, 60, 60)$ and $\Omega_{\text{UP}} = (15, 15, 15)$. The smaller domain represents a scenario in which the coarse sample contains only a small number of fractures with limited influence on the overall hydraulic conductivity.
Given a homogenization block size of $l = 15$ $(1.5H)$, with an overlap of $l/2$, the number of homogenization blocks per sample is $729$ for $\Omega_{\text{UP}} = (60, 60, 60)$ and $27$ for $\Omega_{\text{UP}} = (15, 15, 15)$, respectively.

\subsubsection{Constraint problem}
We first compare the upscaled DFM models for the Constraint problem described in Section~\ref{constraint_problem_3D}.
Table~\ref{tab:outflow_tar_pred_domain_60} reports the $\text{NRMSE}$ between the target outflow $Y$ (using numerical homogenization for upscaling) and the prediction outflow (upscaling by surrogates), for the domain $\Omega_{\text{UP}} = (60, 60, 60)$.
In all three cases, using surrogates to determine $\tn K_H$ produces coarse model outflow values $Y$ that closely match the reference, with $\text{NRMSE} < 0.01$. Moreover, the absolute values of $Y$ are consistent across all three $K_f/K_m$ ratios, suggesting that remaining fractures predominantly control the outflow. As the influence of fractures increases with $K_f/K_m$, the $\text{NRMSE}$ further decreases, ignoring different surrogates' predictive performance.

\begin{table}
\caption{$\text{NRMSE}$ comparison of Constraint problem outflow $Y$ obtained from coarse DFM models, evaluated for different fracture-to-matrix conductivity ratios. For each ratio, numerical homogenization vs. surrogate-based upscaling effect is compared. Parameters: $H = 10$, $h = 5$, $\Omega_{\text{UP}} = (60, 60, 60)$, and $250$ samples.}

\label{tab:outflow_tar_pred_domain_60}
\begin{tabularx}{0.5\textwidth}{CCC}
   \toprule
       \multirow{2.5}{*} & \multirow{2.5}{*}{$K_f/K_m$} 
        & {$\mathbf{NRMSE}$} \\
         \cmidrule{3-3} 
        & & {$\mathbf{Y}$} \\
        \midrule
       Surrogate A & $\num{1e3}$ & $0.00968$   \\
         \cmidrule(lr){1-3} 
           Surrogate B & $\num{1e5}$ & $0.00418$   \\
         \cmidrule(lr){1-3} 
          Surrogate C  & $\num{1e7}$ & $0.00381$    \\
        \bottomrule
\end{tabularx}
\end{table}

Table~\ref{tab:outflow_tar_pred_domain_15} presents results for the smaller domain, $\Omega_{\text{UP}} = (15, 15, 15)$.
The $\text{NRMSE}$ values for the coarse model outflow $Y$ - comparing numerical homogenization and surrogate-based upscaling - reflect trends in the prediction accuracy of the surrogates on the test data (see Section~\ref{prediction_accuracy_on_test_sets}).
The closest agreement occurs for $K_f/K_m = \num{1e3}$, where Surrogate A also achieved the highest accuracy on the test set. For this smaller domain, the upscaled conductivities have a more pronounced impact on the outflow $Y$, resulting in generally worse $\text{NRMSE}$ values compared to the larger domain case $\Omega_{\text{UP}} = (60, 60, 60)$.

\begin{table}
\caption{$\text{NRMSE}$ comparison of the coarse Constraint problem outflow $Y$, evaluated for different fracture-to-matrix conductivity ratios. For each ratio, numerical homogenization vs. surrogate-based upscaling effect is compared. Parameters: $H = 10$, $h = 5$, $\Omega_{\text{UP}} = (15, 15, 15)$, and $250$ samples.}
\label{tab:outflow_tar_pred_domain_15}
\begin{tabularx}{0.5\textwidth}{CCC}
   \toprule
       \multirow{2.5}{*} & \multirow{2.5}{*}{$K_f/K_m$} 
        & {$\mathbf{NRMSE}$} \\
         \cmidrule{3-3} 
        & & {$\mathbf{Y}$} \\
        \midrule
       Surrogate A & $\num{1e3}$   & $0.01323$   \\
         \cmidrule(lr){1-3} 
           Surrogate B & $\num{1e5}$   & $0.06235$   \\
         \cmidrule(lr){1-3} 
          Surrogate C  & $\num{1e7}$ & $0.06419$    \\
        \bottomrule
\end{tabularx}
\end{table}

\subsubsection{Anisotropy problem}
Second, we compare the equivalent hydraulic conductivity tensors of the coarse DFM models, upscaled by numerical homogenization and by the surrogates. 
The equivalent hydraulic conductivity tensor $\tn K^{eq}_{\text{UP}}$, is calculated as described in Section~\ref{anisotropy_problem_3D}, but for the entire domain $\Omega_{\text{UP}}$.
As shown in Table ~\ref{tab:equivalent_tensor_accuracy_domain_60}, the $\text{NRMSE}$ values for $\Omega_{\text{UP}} = (60, 60, 60)$ case (see Table) demonstrate that all surrogates provide highly accurate predictions of $\tn K^{eq}$, regardless of the value of $K_f/K_m \in \{\num{1e3}, \num{1e5}, \num{1e7}\}$ used. 
Notably, the overall $\text{NRMSE}$ is an order of magnitude lower than in the Constraint problem. This observation aligns with findings from the 2D case (\cite{Spetlik2024}), where the Constraint problem exhibits greater sensitivity to matrix conductivity values than the Anisotropy problem.

\begin{table}
\caption{
$\text{NRMSE}$ comparison of the components of the equivalent hydraulic conductivity tensor $\tn K^{eq}_\text{UP}$ obtained from the coarse DFM models for different fracture-to-matrix conductivity ratios. Surrogate-based upscaling is compared against numerical homogenization for each ratio. Parameters: $H = 10$, $h = 5$, domain size $\Omega_{\text{UP}} = (60, 60, 60)$, and $250$ samples.
}
\label{tab:equivalent_tensor_accuracy_domain_60}
\begin{tabularx}{\textwidth}{CCCCCCCC}
   \toprule
       \multirow{2.5}{*} &  \multirow{2.5}{*}{$K_f/K_m$} 
       & \multicolumn{6}{c}{$\mathbf{NRMSE}$} \\
       \cmidrule{3-8} 
       & & $\mathbf{k_{xx}}$& $\mathbf{k_{yy}}$& $\mathbf{k_{zz}}$ & $\mathbf{k_{yz}}$ & $\mathbf{k_{xz}}$ & $\mathbf{k_{xy}}$ \\
        \midrule
       Surrogate A & $\num{1e3}$   & $0.00133$ & $0.00054$ & $0.00046$ & $0.00057$ & $0.00084$ & $0.00053$   \\
         \cmidrule(lr){1-8} 
           Surrogate B & $\num{1e5}$   & $0.00038$ & $0.00022$ & $0.00030$ & $0.00064$ & $0.00037$ & $0.00065$   \\
         \cmidrule(lr){1-8} 
          Surrogate C  & $\num{1e7}$ & $0.00022$ & $0.00022$ & $0.00033$ & $0.00041$ & $0.00022$ & $0.00025$   \\
        \bottomrule
\end{tabularx}
\end{table}

\begin{table}
\caption{
$\text{NRMSE}$ comparison of the components of the equivalent hydraulic conductivity tensor $\tn K^{eq}_{\text{UP}}$ obtained from the coarse DFM models for different fracture-to-matrix conductivity ratios. Surrogate-based upscaling is compared against numerical homogenization for each ratio. Parameters: $H = 10$, $h = 5$, domain size $\Omega_{\text{UP}} = (15, 15, 15)$, and $250$ samples.
}

\label{tab:equivalent_tensor_accuracy_domain_15}
\begin{tabularx}{\textwidth}{CCCCCCCC}
   \toprule
        \multirow{2.5}{*} &  \multirow{2.5}{*}{$K_f/K_m$} 
       & \multicolumn{6}{c}{$\mathbf{NRMSE}$} \\
       \cmidrule{3-8} 
       & & $\mathbf{k_{xx}}$& $\mathbf{k_{yy}}$& $\mathbf{k_{zz}}$ & $\mathbf{k_{yz}}$ & $\mathbf{k_{xz}}$ & $\mathbf{k_{xy}}$ \\
        \midrule
       Surrogate A & $\num{1e3}$   & $0.01222$ & $0.01372$ & $0.01793$ & $0.03631$ & $0.04027$ & $0.03262$   \\
         \cmidrule(lr){1-8} 
           Surrogate B & $\num{1e5}$   & $0.01483$ & $0.01297$ & $0.01289$ & $0.02076$ & $0.01229$ & $0.01897$   \\
         \cmidrule(lr){1-8} 
          Surrogate C  & $\num{1e7}$ & $0.01543$ & $0.01228$ & $0.01247$ & $0.01562$ & $0.01272$ & $0.01179$   \\
        \bottomrule
\end{tabularx}
\end{table}

As shown in Table~\ref{tab:equivalent_tensor_accuracy_domain_15}, the Anisotropy problem also reveals a decline in the accuracy of the $\tn K^{eq}_{\text{UP}}$ components when the influence of fractures resolved by the coarse model is limited, as in the smaller domain $\Omega_{\text{UP}} = (15, 15, 15)$.
Interestingly, the accuracy of the off-diagonal components improves with increasing $K_f/K_m$, which can be attributed to the decreasing relative importance of surrogate-predicted matrix values as $K_f/K_m$ increases, leading to reduced sensitivity to prediction errors.

For both problems presented, we observe that even in cases where the influence of the upscaled conductivity is significant, the trained surrogates produce results that remain in close agreement with those obtained via standard numerical homogenization.

\section{Conclusions}\label{conclussion_section}
This study introduced a deep learning surrogate designed to predict the equivalent hydraulic conductivity tensor $\tn K^{eq}$ from 3D discrete fracture-matrix (DFM) models. We evaluated the surrogate's predictive performance as well as its utility in macro-scale (coarse model) hydrogeological simulations.
Three datasets were formed corresponding to fracture-to-matrix hydraulic conductivity ratios of $K_f/K_m \in \{\num{1e3}, \num{1e5}, \num{1e7}\}$. Surrogates trained on these datasets achieved high predictive accuracy on the test dataset, with $\text{NRMSE} < 0.22$ for all tensor components. 

We assessed the generalizability of the surrogates across different fracture network configurations and spatial correlation lengths of the matrix random field. 
Higher fracture densities, beyond those seen in training, resulted in a decline in performance. When tested on samples from four distinct fracture network configurations, the trained surrogates exhibited only a slight reduction in predictive accuracy.
The impact of changes in spatial correlation length depended on the value of $K_f/K_m$: prediction accuracy remained consistent for correlation lengths within the training range but deteriorated for significantly larger values.

The surrogates demonstrated substantial computational advantages across various numbers of homogenization blocks in use. The upscaling with the surrogate was approximately $100\times$ faster than standard numerical homogenization when GPU-based inference was employed. 

To evaluate the impact of surrogate accuracy, we embedded the upscaled $\tn K^{eq}$ - obtained via numerical homogenization and surrogates - into two macro-scale hydrological problems. In both the Constraint and Anisotropy problems, the quantities of interest were only marginally affected by surrogate-based upscaling, even in scenarios where the influence of the upscaled conductivities was substantial. The Constraint problem exhibited greater sensitivity to the bulk conductivities and, consequently, to the accuracy of the surrogate predictions.

Overall, the proposed 3D surrogate models offer a scalable and effective tool for upscaling in fractured porous media, with promising generalizability across a broad range of settings. 
Future work will focus on integrating these surrogates into multilevel Monte Carlo frameworks, including investigating the impact of the homogenization block size.

\section{Acknowledgments}
The research was supported by the European Partnership on Radioactive Waste Management 2 (EURAD-2). EURAD-2 is co-funded by the European Union under Grant Agreement N° 101166718.
This work was (partly) supported by the Student Grant Scheme at the Technical University of Liberec through project nr. SGS-2025-3560.
Computational resources were provided by the e-INFRA CZ project (ID:90254),
supported by the Ministry of Education, Youth and Sports of the Czech Republic.
We acknowledge VSB – Technical University of Ostrava, IT4Innovations National Supercomputing Center, Czech Republic, for awarding this project access to the LUMI supercomputer, owned by the EuroHPC Joint Undertaking, hosted by CSC (Finland) and the LUMI consortium through the Ministry of Education, Youth and Sports of the Czech Republic through the e-INFRA CZ (grant ID: 90254).


\newpage

\textbf{Code availability section}
The source codes are available for downloading at the link:
\url{https://github.com/martinspetlik/MLMC-DFM/tree/MS_3d}

\bibliographystyle{cas-model2-names}
\bibliography{bibliography_revised} 

@article{Giles2015,
  title = {Multilevel {{{M}onte {C}arlo}} Methods},
  volume = {24},
  issn = {0962-4929, 1474-0508},
  doi = {10.1017/S096249291500001X},
  language = {en},
  journal = {Acta Numerica},
  author = {Giles, Michael B.},
  year = {2015},
  pages = {259-328},
}

@book{Sahimi20110420,
  author = {Muhammad Sahimi},
  title = {Flow and Transport in Porous Media and Fractured Rock},
  publisher = {Wiley},
  year = {2011},
  edition = {2.},
  ISBN = {9783527404858},
  medium = {online},
  DOI = {10.1002/9783527636693},
}

@article{Lang20140815,
  author = {P. S. Lang and A. Paluszny and R. W. Zimmerman},
  journal = {Journal of Geophysical Research: Solid Earth},
  title = {Permeability tensor of three‐dimensional fractured porous rock and a comparison to trace map predictions},
  year = {2014},
  pages = {6288-6307},
  volume = {119},
  number = {8},
  ISSN = {2169-9313},
  medium = {online},
  DOI = {10.1002/2014JB011027},
  URL = {https://onlinelibrary.wiley.com/doi/abs/10.1002/2014JB011027},
}

@book{Auriault20090101,
  author = {Jean-Louis Auriault and Claude Boutin and Christian Geindreau},
  title = {Homogenization of Coupled Phenomena in Heterogenous Media},
  publisher = {ISTE},
  address = {London, UK},
  year = {2009},
  ISBN = {9780470612033},
  medium = {online},
  DOI = {10.1002/9780470612033},
}

@book{Banks2002,
  author = {David Banks and Nick Robins},
  title = {An introduction to groundwater in crystalline bedrock},
  publisher = {Geological Survey of Norway},
  year = {2002},
  ISBN = {82-7386-100-1},
}

@article{PhysRevE.76.036309,
  title = {Effective permeability of fractured porous media with power-law distribution of fracture sizes},
  author = {Bogdanov, I. I. and Mourzenko, V. V. and Thovert, J.-F. and Adler, P. M.},
  journal = {Phys. Rev. E},
  volume = {76},
  issue = {3},
  pages = {036309},
  numpages = {15},
  year = {2007},
  month = {Sep},
  publisher = {American Physical Society},
  doi = {10.1103/PhysRevE.76.036309},
  url = {https://link.aps.org/doi/10.1103/PhysRevE.76.036309}
}

@article{VASILYEVA2021185,
title = {Machine learning for accelerating macroscopic parameters prediction for poroelasticity problem in stochastic media},
journal = {Computers \& Mathematics with Applications},
volume = {84},
pages = {185-202},
year = {2021},
issn = {0898-1221},
doi = {https://doi.org/10.1016/j.camwa.2020.09.024},
url = {https://www.sciencedirect.com/science/article/pii/S0898122121000067},
author = {Maria Vasilyeva and Aleksey Tyrylgin}
}

@misc{bgem,
  author       = {Jan Březina and Jiří Kopal and Radek Srb and Jana Ehlerová and Jiří Hnídek},
  title        = {bgem: B-spline GEometry Modeling package},
  version      = {0.3.0},
  year         = 2025,
  url          = {https://github.com/GeoMop/bgem},
}

@article{Spetlik2024,
  title     = "Deep learning surrogate for predicting hydraulic conductivity
               tensors from stochastic discrete fracture-matrix models",
  author    = "{\v S}petl{\'\i}k, Martin and B{\v r}ezina, Jan and Laloy, Eric",
  journal   = "Computational Geosciences",
  publisher = "Springer Science and Business Media LLC",
  month     =  oct,
  year      =  2024,
  copyright = "https://www.springernature.com/gp/researchers/text-and-data-mining",
  language  = "en",
doi = {https://doi.org/10.1007/s10596-024-10324-8},
url = {https://link.springer.com/article/10.1007/s10596-024-10324-8},
}

@techreport{SKB-R-09-20,
  key          = {SKB R-09-20},
  title        = {Groundwater Flow Modelling of Periods with Temperate Climate Conditions \textendash{} Forsmark},
  institution  = {Svensk Kärnbränslehantering AB},
  number       = {SKB R-09-20},
  language     = {english},
  author       = {Joyce, Steven and Simpson, Trevor and Hartley, Lee and Applegate, David and Hoek, Jaap and Jackson, Peter and Swan, David and Marsic, Niko and Follin, Sven},
  pages        = {315},
  year         = {2009}
}

@techreport{SKB-R-08-78,
  key         = {SKB R-08-78},
  title       = {Hydrogeological Conceptualisation and Parameterization, Site Descriptive Modelling SDM-Site Laxemar},
  number      = {SKB R-08-78},
  institution = {Svensk Kärnbränslehantering AB},
  language    = {english},
  author      = {Rhén, Ingvar and Forsmark, Torbjörn and Hartley, Lee and Jackson, Peter and Roberts, David and Swan, Dave and Gylling, Björn},
  pages       = {745},
  year        = {2008}
}

@article{Hadgu_comparative_2017,
  title = {A Comparative Study of Discrete Fracture Network and Equivalent Continuum Models for Simulating Flow and Transport in the Far Field of a Hypothetical Nuclear Waste Repository in Crystalline Host Rock},
  author = {Hadgu, Teklu and Karra, Satish and Kalinina, Elena and Makedonska, Nataliia and Hyman, Jeffrey D. and Klise, Katherine and Viswanathan, Hari S. and Wang, Yifeng},
  year = {2017},
  journal = {Journal of Hydrology},
  volume = {553},
  pages = {59--70},
  issn = {0022-1694},
  doi = {10.1016/j.jhydrol.2017.07.046},
  urldate = {2024-02-26}
}

@article{Kottwitz_Investigating_2021,
  title = {Investigating the Effects of Intersection Flow Localization in Equivalent-Continuum-Based Upscaling of Flow in Discrete Fracture Networks},
  author = {Kottwitz, Maximilian O. and Popov, Anton A. and Abe, Steffen and Kaus, Boris J. P.},
  year = {2021},
  journal = {Solid Earth},
  volume = {12},
  number = {10},
  pages = {2235--2254},
  publisher = {{Copernicus GmbH}},
  issn = {1869-9510},
  doi = {10.5194/se-12-2235-2021},
  urldate = {2024-02-26},
  langid = {english}
}

@Misc{flow123d,
author =   {B{\v r}ezina, Jan and Stebel, Jan and Exner, Pavel and Hyb{\v s}, Jan},
title =    {Flow123d},
howpublished = {\url{http://flow123d.github.com}, repository: \url{http://github.com/flow123d/flow123d}},
year = {2011--2025}
}

@Misc{GSTools,
author =   {Müller, Sebastian and Schüler, Lennart},
title =    {{GSTools}},
howpublished = {\url{https://github.com/GeoStat-Framework/GSTools}},
year = {2019}
}

@book{Goodfellow-et-al-2016,
    title={Deep Learning},
    author={Ian Goodfellow and Yoshua Bengio and Aaron Courville},
    publisher={MIT Press},
    address={Cambridge, Massachusetts},
    note={\url{http://www.deeplearningbook.org}},
    year={2016}
}

@article{Bonnet2001ScalingOF,
author = {Bonnet, E. and Bour, O. and Odling, N. E. and Davy, P. and Main, I. and Cowie, P. and Berkowitz, B.},
title = {Scaling of fracture systems in geological media},
journal = {Reviews of Geophysics},
volume = {39},
number = {3},
pages = {347-383},
doi = {https://doi.org/10.1029/1999RG000074},
year = {2001}
}

@book{Adler1999,
  author = {Pierre M. Adler and Jean-François Thovert},
  title = {Fractures and Fracture Networks},
  publisher = {Springer Netherlands},
  address = {Dordrecht},
  year = {1999},
  ISBN = {978-90-481-5192-9},
  DOI = {10.1007/978-94-017-1599-7},
}

@article{Azizmohammadi2017,
  author = {Siroos Azizmohammadi and Stephan K. Matthäi},
  journal = {Water Resources Research},
  title = {Is the permeability of naturally fractured rocks scale dependent?},
  year = {2017},
  pages = {8041-8063},
  volume = {53},
  number = {9},
  ISSN = {00431397},
  medium = {online},
  DOI = {10.1002/2016WR019764},
  URL = {http://doi.wiley.com/10.1002/2016WR019764},
}

@article{RAO2020109850,
title = {Three-dimensional convolutional neural network (3D-CNN) for heterogeneous material homogenization},
journal = {Computational Materials Science},
volume = {184},
pages = {109850},
year = {2020},
issn = {0927-0256},
doi = {https://doi.org/10.1016/j.commatsci.2020.109850},
url = {https://www.sciencedirect.com/science/article/pii/S0927025620303414},
author = {Chengping Rao and Yang Liu}
}

@article{Berrone_Simulations_2013,
  title = {On {{Simulations}} of {{Discrete Fracture Network Flows}} with an {{Optimization-Based Extended Finite Element Method}}},
  author = {Berrone, Stefano and Pieraccini, Sandra and Scial{\`o}, Stefano},
  year = {2013},
  journal = {SIAM Journal on Scientific Computing},
  volume = {35},
  number = {2},
  pages = {A908-A935},
  publisher = {{Society for Industrial and Applied Mathematics}},
  issn = {1064-8275},
  doi = {10.1137/120882883},
  urldate = {2024-03-18}
}

@article{Sandve_efficient_2012a,
  title = {An Efficient Multi-Point Flux Approximation Method for {{Discrete Fracture}}--{{Matrix}} Simulations},
  author = {Sandve, T. H. and Berre, I. and Nordbotten, J. M.},
  year = {2012},
  journal = {Journal of Computational Physics},
  volume = {231},
  number = {9},
  pages = {3784--3800},
  issn = {0021-9991},
  doi = {10.1016/j.jcp.2012.01.023},
  urldate = {2024-03-18}
}

@InProceedings{Brezina2016Analysis,
author="B{\v{r}}ezina, Jan
and Stebel, Jan",
editor="Kozubek, Tom{\'a}{\v{s}}
and Blaheta, Radim
and {\v{S}}{\'i}stek, Jakub
and Rozlo{\v{z}}n{\'i}k, Miroslav
and {\v{C}}erm{\'a}k, Martin",
title="Analysis of Model Error for a Continuum-Fracture Model of Porous Media Flow",
booktitle="High Performance Computing in Science and Engineering",
year="2016",
publisher="Springer International Publishing",
doi = {10.1007/978-3-319-40361-8_11},
address="Cham",
pages="152--160",
isbn="978-3-319-40361-8"
}

@article{Berre2019,
  author = {Inga Berre and Florian Doster and Eirik Keilegavlen},
  journal = {Transport in Porous Media},
  title = {Flow in Fractured Porous Media},
  subtitle = {A Review of Conceptual Models and Discretization Approaches},
  year = {2019},
  pages = {215-236},
  volume = {130},
  number = {1},
  ISSN = {0169-3913},
  medium = {online},
  DOI = {10.1007/s11242-018-1171-6},
  URL = {http://link.springer.com/10.1007/s11242-018-1171-6},
}

@article{FERREIRA2022104264,
title = {A framework for upscaling and modelling fluid flow for discrete fractures using conditional generative adversarial networks},
journal = {Advances in Water Resources},
volume = {166},
pages = {104264},
year = {2022},
issn = {0309-1708},
doi = {https://doi.org/10.1016/j.advwatres.2022.104264},
url = {https://www.sciencedirect.com/science/article/pii/S0309170822001336},
author = {Carlos A.S. Ferreira and Teeratorn Kadeethum and Nikolaos Bouklas and Hamidreza M. Nick}
}

@article{https://doi.org/10.1029/2001WR000756,
author = {Bogdanov, I. I. and Mourzenko, V. V. and Thovert, J.-F. and Adler, P. M.},
title = {Effective permeability of fractured porous media in steady state flow},
journal = {Water Resources Research},
volume = {39},
number = {1},
pages = {},
doi = {https://doi.org/10.1029/2001WR000756},
url = {https://agupubs.onlinelibrary.wiley.com/doi/abs/10.1029/2001WR000756},
year = {2003}
}

@article{Hong2020,
  author = {Jin Hong and Jie Liu},
  journal = {Computational Geosciences},
  title = {Rapid estimation of permeability from digital rock using 3D convolutional neural network},
  year = {2020},
  pages = {1523-1539},
  volume = {24},
  number = {4},
  ISSN = {1420-0597},
  medium = {online},
  DOI = {10.1007/s10596-020-09941-w},
  URL = {https://link.springer.com/10.1007/s10596-020-09941-w},
}

@article{Cai_2023,
	author  = {Chen  Cai and Nikolaos  Vlassis and Lucas  Magee and Ran  Ma and Zeyu  Xiong and Bahador  Bahmani and Teng-Fong  Wong and Yusu  Wang and WaiChing Sun},
	title   = {Equivariant geometric learning for digital rock physics: estimating
formation factor and effective permeability tensors from Morse graph},
	journal = {International Journal for Multiscale Computational Engineering},
	issn    = {1543-1649},
	year    = {2023},
	volume  = {21},
	number  = {5},
	pages   = {1--24}
}

@Article{pr11020601,
AUTHOR = {Pal, Mayur and Makauskas, Pijus and Malik, Shruti},
TITLE = {Upscaling Porous Media Using Neural Networks: A Deep Learning Approach to Homogenization and Averaging},
JOURNAL = {Processes},
VOLUME = {11},
YEAR = {2023},
NUMBER = {2},
ARTICLE-NUMBER = {601},
URL = {https://www.mdpi.com/2227-9717/11/2/601},
ISSN = {2227-9717},
DOI = {10.3390/pr11020601}
}

@article{MENG2023104520,
title = {Transformer-based deep learning models for predicting permeability of porous media},
journal = {Advances in Water Resources},
volume = {179},
pages = {104520},
year = {2023},
issn = {0309-1708},
doi = {https://doi.org/10.1016/j.advwatres.2023.104520},
url = {https://www.sciencedirect.com/science/article/pii/S0309170823001549},
author = {Yinquan Meng and Jianguo Jiang and Jichun Wu and Dong Wang}
}

@article{STEPANOV2023114980,
title = {Prediction of numerical homogenization using deep learning for the Richards equation},
journal = {Journal of Computational and Applied Mathematics},
volume = {424},
pages = {114980},
year = {2023},
issn = {0377-0427},
doi = {https://doi.org/10.1016/j.cam.2022.114980},
url = {https://www.sciencedirect.com/science/article/pii/S0377042722005787},
author = {Sergei Stepanov and Denis Spiridonov and Tina Mai}
}

@inproceedings{10.2118/203901-MS,
    author = {He, Xupeng  and Santoso, Ryan  and Alsinan, Marwa  and Kwak, Hyung  and Hoteit, Hussein },
    title = "{Constructing Dual-Porosity Models from High-Resolution Discrete-Fracture Models Using Deep Neural Networks}",
    volume = {Day 1 Tue, October 26, 2021},
    series = {SPE Reservoir Simulation Conference},
    pages = {D011S014R012},
    year = {2021},
    month = {10},
    booktitle = {Proceedings of the SPE Reservoir Simulation Conference},
    doi = {10.2118/203901-MS},
}

@article{Andrianov2022UpscalingOT,
  title={Upscaling of two-phase discrete fracture simulations using a convolutional neural network},
  author={Nikolai Andrianov},
  journal={Computational Geosciences},
  year={2022},
  volume={26},
  pages={1237 - 1259},
  doi = {https://doi.org/10.1007/s10596-022-10149-3},
  url={https://api.semanticscholar.org/CorpusID:249551231}
}

@article{Wang2023,
  author = {Nanzhe Wang and Qinzhuo Liao and Haibin Chang and Dongxiao Zhang},
  journal = {Computational Geosciences},
  title = {Deep-learning-based upscaling method for geologic models via theory-guided convolutional neural network},
  year = {2023},
  pages = {913-938},
  volume = {27},
  number = {6},
  ISSN = {1420-0597},
  accessed = {2024-04-25},
  DOI = {10.1007/s10596-023-10233-2},
  URL = {https://link.springer.com/10.1007/s10596-023-10233-2},
}

@book{IAEA_2011_SSG14,
  address   = {Vienna},
  edition   = {1},
  editor    = {{IAEA}},
  isbn      = {978-92-0-111510-2},
  pages     = {104},
  publisher = {International Atomic Energy Agency},
  series    = {IAEA Safety Standards Series: Specific Safety Guide No. SSG-14},
  title     = {Geological Disposal Facilities for Radioactive Waste},
  year      = {2011},
    URL = {https://www-pub.iaea.org/MTCD/Publications/PDF/Pub1483_web.pdf}
}

@article{WEN1996ix,
title = {Upscaling hydraulic conductivities in heterogeneous media: An overview},
journal = {Journal of Hydrology},
volume = {183},
number = {1},
pages = {ix-xxxii},
year = {1996},
issn = {0022-1694},
doi = {https://doi.org/10.1016/S0022-1694(96)80030-8},
url = {https://www.sciencedirect.com/science/article/pii/S0022169496800308},
author = {Xian-Huan Wen and J.Jaime Gómez-Hernández},
}

\end{document}